\documentclass[a4paper,fleqn]{cas-sc}

\usepackage[numbers]{natbib}
\usepackage{booktabs}    
\usepackage{ulem}
\usepackage{array, caption, threeparttable}
\usepackage[font=small,labelfont=bf,labelsep=none]{caption}
\usepackage{amsmath}
\def\tsc#1{\csdef{#1}{\textsc{\lowercase{#1}}\xspace}}
\tsc{WGM}
\tsc{QE}
\tsc{EP}
\tsc{PMS}
\tsc{BEC}
\tsc{DE}

\begin{document}
\let\WriteBookmarks\relax
\def\floatpagepagefraction{1}
\def\textpagefraction{.001}

\shorttitle{Leveraging social media news}

\shortauthors{Maonian Wu et~al.}

\title [mode = title]{Neuro-Symbolic Recommendation Model based on Logic Query}                      



%
\author[1,2]{Maonian Wu}[type=editor,
                        style=chinese,
                        auid=000,bioid=1,
                        ]


\ead{wmn@zjhu.edu.cn}


\credit{Conceptualization, Writing – original draft, Writing – review \& editing}

\author[1,2]{Bang Chen}[style=chinese,
                        orcid=0000-0003-1493-9871
                        ]
\credit{Conceptualization, Methodology, Software, Data curation, Writing – original draft}
\ead{2021388105@stu.zjhu.edu.cn}
\author[1,2]{Shaojun Zhu}[style=chinese]
\credit{Resources}

\author[1,2]{Bo Zheng}[style=chinese]
\credit{Funding acquisition}
\author[3]{Wei Peng}[style=chinese]
\credit{Writing - Review \& Editing}

\author[4]{Mingyi Zhang}[style=chinese,
                        ]
\cormark[1]
\credit{Validation, Supervision, Writing - Review \& Editing}
\ead{zhangmingyi045@outlook.com}

\affiliation[1]{organization={College of Computer Science},
    addressline={Huzhou University}, 
    city={Huzhou},
    country={China}}
\affiliation[2]{organization={Zhejiang Province Key
Laboratory of Smart Management and Application of Modern Agricultural Resources},
    city={Huzhou},
    country={China}}
\affiliation[3]{organization={College of Computer Science},
    addressline={Guizhou University}, 
    city={Guiyang},
    country={China}}
\affiliation[4]{organization={Guizhou Academy of Sciences},
    city={Guiyang},
    country={China}}

\cortext[cor1]{Corresponding author}

\begin{abstract}
A recommendation system assists users in finding items that are relevant to them. Existing recommendation models are primarily based on predicting relationships between users and items and use complex matching models or incorporate extensive external information to capture association patterns in data. However, recommendation is not only a problem of inductive statistics using data; it is also a cognitive task of reasoning decisions based on knowledge extracted from information. Hence, a logic system could naturally be incorporated for the reasoning in a recommendation task. However, although hard-rule approaches based on logic systems can provide powerful reasoning ability, they struggle to cope with inconsistent and incomplete knowledge in real-world tasks, especially for complex tasks such as recommendation. Therefore, in this paper, we propose a neuro-symbolic recommendation model, which transforms the user history interactions into a logic expression and then transforms the recommendation prediction into a query task based on this logic expression. The logic expressions are then computed based on the modular logic operations of the neural network. We also construct an implicit logic encoder to reasonably reduce the complexity of the logic computation. Finally, a user's interest items can be queried in the vector space based on the computation results. Experiments on three well-known datasets verified that our method performs better compared to state of the art shallow, deep, session, and reasoning models.
\end{abstract}


\begin{highlights}
\item We transforms recommendation problem into a first-order logic-based query problem.
\item We improve model's characterization ability while reduce its computational complexity.
\item We use separate neural networks for logic predicate operations.
\item We demonstrate our method on several real recommendation datasets.
\end{highlights}

\maketitle

\section{Introduction}
Rapidly developing technology and mobile device upgrades have yielded rich and colorful lifestyles. The further development of e-commerce and new retail provides a variety of retail options for consumers. However, the explosive growth of information and data also presents a problem for consumers: identifying potentially interesting products from tens or even hundreds of millions of products is difficult. Recommendation systems, which can match consumers with the products that are most relevant to them and help them make decisions, have become an important tool to alleviate information overload and improve consumer experience \cite{cui2020personalized}. Hence, in recent years, recommendation systems have become a research hotspot in the field of artificial intelligence.

Most existing recommendation methods are based on collaborative filtering (CF) \cite{schafer2007collaborative}, which is an important method to improve and optimize the model of a recommendation system. Classical CF methods include content-based CF \cite{sarwar2001item} and user-based CF \cite{resnick1994grouplens}, which identify items or users, respectively, with similar characteristics to assist prediction. Some researchers have proposed matrix decomposition algorithms \cite{koren2009matrix,he2016fast}, which abstract the recommendation problem into the problem of decomposing the scoring matrix into a user and an item matrix, where each row in the user matrix represents the characteristics of a user, each column in the item matrix represents the characteristics of an item, and a corresponding matching score is obtained by the dot product of these two vectors. Many scholars have since improved the recommendation performance by identifying better matching functions, such as using Euclidean distance \cite{tay2018latent} or nonlinear neural networks \cite{yin2019deeper,xue2017deep} instead of vector dot products to measure user preferences in the potential space. Some studies have also explored embedding learning, such as bringing in path embedding to enhance global feature representation \cite{tian2022exploiting}, adding more feature embedding information based on temporal sequences \cite{li2017neural,heidari2022attention,tang2022time}, or using richer external heterogeneous information sources, such as sentiment space context \cite{zhao2020personalized} or knowledge graphs \cite{zhang2016collaborative,wang2022tkgat}. Recently, with the rapid development and widespread use of deep learning, using deep models to learn from large-scale data to fit matching functions or fuse more feature embeddings has become a popular research trend \cite{cheng2016wide}.

However, the recommendation problem also requires cognitive reasoning ability \cite{marcus2020next, chen2021neural}. From some perspectives, recommendation prediction is about simulating user decisions based on known information from the past. A growing number of researchers have found that the recommendation problem cannot be perfectly solved by simply performing statistical induction on known data \cite{ferrari2019we,rendle2020neural,bengio2019system}. For example, an individual who has recently bought a computer will no longer need recommendations for similar products, but for peripheral products such as keyboards and mice. However, current recommendation systems often recommend similar products just after an item has been bought, even though the demand has disappeared. Furthermore, statistical matching-based approaches struggle to learn enough user-item feature pairs when datasets are too sparse; the number of possible item-user pairs increases exponentially with the growth of the dataset, and neural network models lack the ability to reason from one to the other and heavily rely on the merits of the dataset. Despite this, few studies use logic systems to solve recommendation problems; this is mainly because hard-rule-based logic approaches often lack generalization, which leads to the inability to accommodate internal conflicts for real-world problems. For example, consumers with the same interaction history may make very different decisions, i.e., there are some conflicting logic rules in the data, which is not allowed in such logic systems.

Recent research advances in neuro-symbolic computing may help bridge the gap between statistical learning and logic reasoning to solve the above problem \cite{harmelen2022preface}. Neuro-symbolic computing attempts to combine advances in connectionist architectures (e.g., artificial neural networks) with symbolic reasoning based on logic systems to build hybrid AI architectures with both generalization and reasoning capabilities \cite{de2020statistical,van2022analyzing}. Inspired by this, we propose a logic query-based neural symbolic recommendation model that transforms a recommendation task into a query problem based on first-order logic. First, a user's historical interaction information is transformed into a first-order predicate logic expression. Then, the solution vector of the logic expression is obtained using a neural network to modulate the logic operations. Finally, the recommendation prediction is obtained by querying items in the vector space. This approach allows the model to benefit from both the formal reasoning of first-order logic and the generalisability of neural networks, addressing problems with current recommender systems, such as dependence on the quality of the dataset, lack of reasoning power, and difficulty in incorporating logical systems. The main contributions of this paper are as follows:
\begin{enumerate}
\itemsep=0pt
\item We propose a reasoning-based neuro-symbolic recommendation model, which transforms a recommendation problem into a first-order logic-based query problem.
\item We add an implicit logic encoder to the model to improve its characterization ability while effectively reducing its computational complexity.
\item We use a separate neural network for logic predicate operations rather than embedding relations as vectors to perform computations.
\item We conduct experiments on several real recommendation datasets and demonstrate that our approach significantly outperforms state-of-the-art recommendation algorithms.
\end{enumerate}
The remainder of this paper is structured as follows. Section 2 presents related work. In Section 3, the flow and framework of the model are described and how the recommendation problem is transformed into a logic expression-based query is detailed. In Section 4, the experimental settings are detailed and the experimental results are analyzed. The paper is concluded in Section 5.

\section{Related work}
The most prevalent approach in recommendation systems is based on the idea of matching, which predicts the relationship between users and items by learning a matching function \cite{huang2019exploring}. The earliest matrix decomposition techniques decompose a user-item interaction matrix into the product of two or more matrices to characterize the relationship between users and items \cite{koren2008factorization}. Currently, researchers optimize recommendation models using three main approaches. The first is to optimize embeddings as the goal. For example, Gediminas et al. \cite{adomavicius2011context} characterized user features using contextual pre-filtering modelling. Alexandros et al. \cite{karatzoglou2010multiverse} used a higher dimensional tensor instead of matrix decomposition to further enrich the embedding information by enhancing the dimensionality. Yehuda \cite{koren2009collaborative} added dynamic temporal information and user behavior to the embedding for learning. The second is to integrate richer information structures in the representations. For example, Zhang et al. \cite{zhang2017joint} used information such as images and ratings to jointly characterize user embeddings. He et al. \cite{he2016vbpr} directly extracted visual features from product images and used them as an independent metric to influence model decisions. Ai et al. \cite{ai2018learning} built recommendation models and optimized the embedding learning with the aid of knowledge graphs. The third optimization method is to build CF recommendation models by finding better-matching functions. For example, Hsieh et al. \cite{hsieh2017collaborative} used vector transformation instead of a vector inner product and measured user preferences in the joint space. As deep learning has shown powerful performance in areas such as images and languages, a growing number of researchers are using complex neural networks to learn matching functions. For example, Ruslan et al. \cite{salakhutdinov2007restricted} first proposed an RBM model combining deep learning with collaborative filtering and achieved good results on a movie dataset. Similar works have also been proposed by He \cite{he2017neural}, who presented a neural collaborative filtering model, and Travis et al. \cite{ebesu2018collaborative}, who presented a collaborative memory network model. All of these approaches use richer prior knowledge or more complex kernel models to enhance recommendation. However, inductive statistics-based approaches struggle to bridge the gap from perception to cognition when faced with tasks that require reasoning power \cite{chen2021neural}. Therefore, the ideas of inductive statistics and reasoning deduction must be integrated to deal with complex recommendation tasks which require some reasoning ability.

Neuro-symbolic computation combines classical symbolic knowledge with neural networks with the aim of providing a better computational approach to integrated machine learning and reasoning. Neuro-symbolic computation enables models to provide both good computational and knowledge reasoning abilities. In the past several decades, researchers have been trying to develop a generalized neuro-symbolic framework; deep learning is seen as a hopeful way to overcome the gap between symbols and sub-symbols for those who want to generate high-level abstract representations from unprocessed low-level data using deep models \cite{jiang2017variational,hohenecker2020ontology,makni2019deep,hinton2006fast}. Recently, deep learning methods have been used to try to solve logic problems; for example, Yang et al. \cite{yang2017differentiable} proposed a neural logic inference system for knowledge repository based on first-order logic. Johnson et al. \cite{johnson2017inferring} and Yi et al. \cite{yi2018neural} both designed deep models to generate programs and perform visual reasoning. Dong et al. \cite{dong2018neural} constructed relational reasoning and decision-making for a neuro-logic model. In addition, some researchers expect to incorporate the idea of symbolic logic in deep models to compensate for the shortcomings of deep learning. For example, Rodríguez et al. \cite{diaz2022explainable} combined expert knowledge graphs with deep learning models to construct an explainable image model. Arabshahi et al. \cite{arabshahi2021conversational} used neuro-symbolic integration models to solve common sense reasoning tasks. Ma et al. \cite{ma2021relvit} used the concept of symbolic manipulation combined with a deep model to implement a visual reasoning model.

At present, neuro-symbolic integration models have made initial progress in many areas, such as images \cite{tsamoura2021neural}, text \cite{lang2021self}, and simple reasoning \cite{amayuelas2022neural}. As a task requiring cognitive ability, recommendation system is a valuable application scenario in which to use a neuro-symbolic approach. The most relevant and inspiring work for us is NLR \cite{shi2020neural}, which transforms a recommendation problem into a logic expression of true or false and solves it by fitting the mapping of propositional logic symbols to rule relations using neural networks. Our work differs in three ways: we construct first-order logic expressions with the help of logic predicates, which better formalize user's historical behavior; we do not compute the true and false values of logic expressions but use them for queries, which is an approach that is more in line with the human decision-making process; and we construct implicit logic encoders to balance the computational complexity of the model with feature mining.

\section{Proposed model}
\begin{figure}[ht]
	\centering
		\includegraphics[scale=.2]{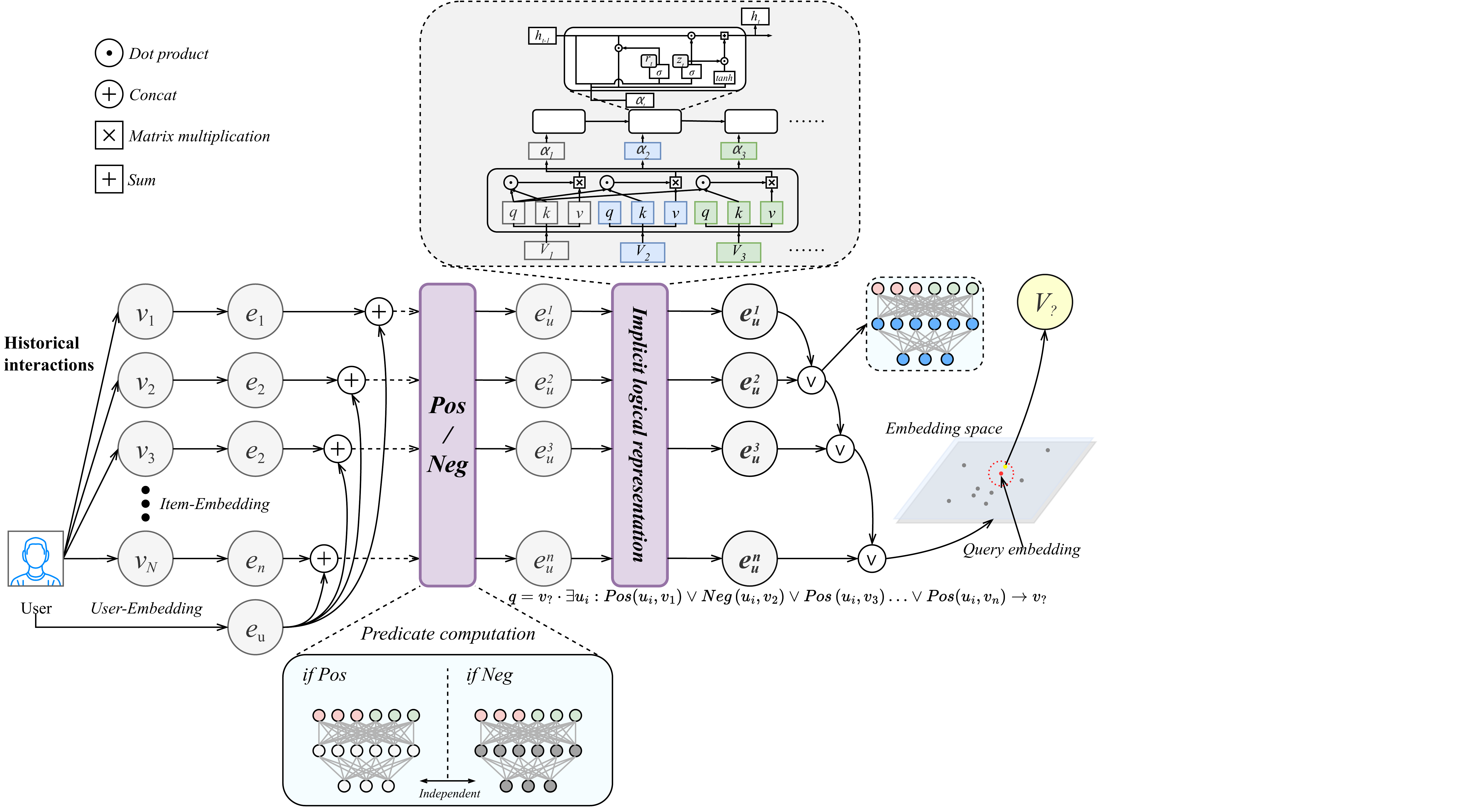}
	\caption{NLR4Rec framework: the proposed model solves the recommendation task from a logic query perspective, using neural networks to compute logic expressions that are transformed from user history interactions and querying target items in a vector space based on the computation results.}
	\label{FIG:1}
\end{figure}

In this section, we first present a statement of the problem and then introduce the proposed neural logic query for recommendation systems, called NLQ4Rec, which aims to transform a recommendation prediction task into a logic query problem and solve it using a neural symbolic approach. Fig. \ref{FIG:1} displays the model's overall structure and will be explained in four parts: (1) logic query-based recommendation; (2) implicit logic encoder; (3) logic operation module; and (4) additional loss function.
\subsection{Problem statement}
For a recommendation task, a model needs to predict a user's future interest items based on their past behaviors. Let $U=\left\{ {{u}_{1}},{{u}_{2}},\ldots ,{{u}_{\left| U \right|}} \right\}$ be a set of all users and $V=\{{{v}_{1}},{{v}_{2}},\ldots ,{{v}_{\left| V \right|}}\}$ be a set of all items; $\left| U \right|$ and $\left| V \right|$ are the total number of users and items in the recommendation task, respectively. Suppose a user ${{u}_{i}}\in U$ has a sequence of historical interactions containing n items, $H=\left\{ {{v}_{1}},{{v}_{2}},\ldots ,{{v}_{n}} \right\}$, where ${{v}_{j}}$ is the $j$th item in the sequence, and a historical rating sequence $R=\left\{ {{r}_{i1}},{{r}_{i2}},\ldots ,{{r}_{in}} \right\}$, where ${{r}_{ij}}$ is user ${{u}_{i}}$'s historical rating of item ${{v}_{j}}$. The goal is to predict which item is most likely to be of interest to a user in the future (i.e., positive interactions) based on their historical interactions and ratings.
\subsection{Recommendation based on logic query}
Traditional recommendation methods usually improve model performance by using a better matching function or considering prior external knowledge; but recommendation is essentially a process that simulates a user's decision-making behavior, and this needs to be driven by cognitive reasoning. To better simulate this decision behavior and perform logic reasoning, we formalized users’ historical interactions as a series of first-order logic formulas. For example, suppose a user ${{u}_{i}}$ has a historical interaction sequence $H=\left\{ {{v}_{1}},{{v}_{2}},{{v}_{3}},{{v}_{4}} \right\}$ and historical rating sequence $R=\left\{ 1,0,1,1 \right\}$; this means that this user has interacted with 4 items, including a negative interaction with item ${{v}_{2}}$ and positive interactions with the other 3 items. This historical information can be broken down into the following logic formulas.
\begin{equation}\label{eq1}
Pos\left( {{u}_{i}},{{v}_{1}} \right)\to Neg\left( {{u}_{i}},{{v}_{2}} \right)
\end{equation}
\vspace*{-1em}
\begin{equation}\label{eq2}
\begin{split}
    Pos({{u}_{i}},{{v}_{1}})\vee Neg\left( {{u}_{i}},{{v}_{2}} \right)\vee \left( Pos({{u}_{i}},{{v}_{1}})\wedge Neg\left( {{u}_{i}},{{v}_{2}} \right) \right) \to Pos({{u}_{i}},{{v}_{3}})
\end{split}
\end{equation}
\vspace*{-1em}
\begin{equation}\label{eq3}
\begin{split}
Pos({{u}_{i}},{{v}_{1}})\vee Neg\left( {{u}_{i}},{{v}_{2}} \right)\vee Pos({{u}_{i}},{{v}_{4}})\vee \left( Pos({{u}_{i}},{{v}_{1}})\wedge Neg\left( {{u}_{i}},{{v}_{2}} \right) \right)\vee \left( Pos({{u}_{i}},{{v}_{1}})\wedge Pos({{u}_{i}},{{v}_{3}}) \right)\vee \\
\left( Neg\left( {{u}_{i}},{{v}_{2}} \right)\wedge Pos({{u}_{i}},{{v}_{3}}) \right)\vee \left( Pos({{u}_{i}},{{v}_{1}})\wedge Neg\left( {{u}_{i}},{{v}_{2}} \right)\wedge Pos({{u}_{i}},{{v}_{3}}) \right)\to Pos({{u}_{i}},{{v}_{4}}) \\ 
\end{split}
\end{equation}
Eq. \ref{eq1} indicates that a user who has a positive interaction with ${{v}_{1}}$ will have a negative interaction with ${{v}_{2}}$. Eq. \ref{eq2} indicates that a user will have a positive interaction with ${{v}_{3}}$ because they either had a positive interaction with ${{v}_{1}}$, a negative interaction with ${{v}_{2}}$, or both. Eq. \ref{eq3} formalizes seven possible rules connected by disjunction symbols. As the recommendation task need not recommend items that the user will not like, we can discard rules that reason negative interactions, such as in Eq. \ref{eq1}, and only keep rules that reason positive interactions; then, we can simplify the term on the right-hand side of the implication symbol. For example, Eq. \ref{eq2} can be simplified to Eq. \ref{eq4}, where each logic expression that represents a user's historical behavior implies an item that the user will like in the future.
\begin{equation}\label{eq4}
\begin{split}
    Pos({{u}_{i}},{{v}_{1}})\vee Neg\left( {{u}_{i}},{{v}_{2}} \right)\vee \left( Pos({{u}_{i}},{{v}_{1}})\wedge Neg\left( {{u}_{i}},{{v}_{2}} \right) \right) \to {{v}_{3}}
\end{split}
\end{equation}
From this, we can translate any known user behavior in recommendation tasks into logic rules and use the model to learn these rules to reason a user's future interest in an item. For example, if user ${{u}_{i}}$ only has a historical interaction sequence $H=\left\{ {{v}_{1}},{{v}_{2}} \right\}$ and historical rating sequence $R=\left\{ 1,0 \right\}$, when the model needs to predict the next item of interest, the problem can be transformed into the following logic query formula based on Eq. \ref{eq4}.
\begin{equation}\label{eq5}
\begin{split}
    q={{v}_{?}}\cdot \exists {{u}_{i}}:Pos({{u}_{i}},{{v}_{1}})\vee Neg\left( {{u}_{i}},{{v}_{2}} \right)\vee \left( Pos({{u}_{i}},{{v}_{1}})\wedge Neg\left( {{u}_{i}},{{v}_{2}} \right) \right)\to {{v}_{?}}
\end{split}
\end{equation}
This is the query: "Find the items that will be of interest to users who interacted positively with ${{v}_{1}}$ and negatively with ${{v}_{2}}$". The answer is ${{v}_{?}}={{v}_{3}}$. Any user history interactions can be transformed into logic query expressions such as Eq. \ref{eq5} and be trained on the model. The logic expressions from all users together form the rule domain and assist the model in providing recommendation predictions.

The recommendation problem has now been initially transformed into a query problem based on logic formulas; details on how to solve and train the model based on logic queries will be presented later. To facilitate the reader's understanding, we first describe the general framework of the model, which is shown in Fig. \ref{FIG:1}. Users and items are first embedded as vectors that can be trained by the neural network. Subsequently, the model constructs a directed acyclic graph (DAG) which defines the order of operations based on a logic expression and uses the neural network to modularize the logic operators to compute this logic expression (detailed in Subsection 3.4). Finally, the most similar items in the item vector space are queried as recommendation predictions based on the solution vector of this expression. Note that the logic query formula in Fig. \ref{FIG:1} is a disjunction paradigm consisting only of binary predicates such as $Pos({{u}_{i}},{{v}_{1}})$,  and does not contain conjunction terms such as $\left( Pos({{u}_{i}},{{v}_{1}})\wedge Neg\left( {{u}_{i}},{{v}_{2}} \right) \right)$ or $\left( Pos({{u}_{i}},{{v}_{1}})\wedge Neg\left( {{u}_{i}},{{v}_{2}} \right)\wedge Pos({{u}_{i}},{{v}_{3}}) \right)$ , which consist of multiple binary predicates. The reason for this is that as the number of historical interactions $n$ grows, the number of disjunction terms in the logic query formula exponentially explodes(e.g., Eq. \ref{eq3}). Therefore, we only consider the disjunction paradigm consisting of binary predicates and construct an implicit logic encoder to capture the higher-order interactions from the logic variables to reach a balance between the computational complexity and accuracy of the model, which we explain in more detail in Subsection 3.3.
\subsection{Implicit logic encoder}
\begin{figure}[ht]
	\centering
		\includegraphics[scale=.15]{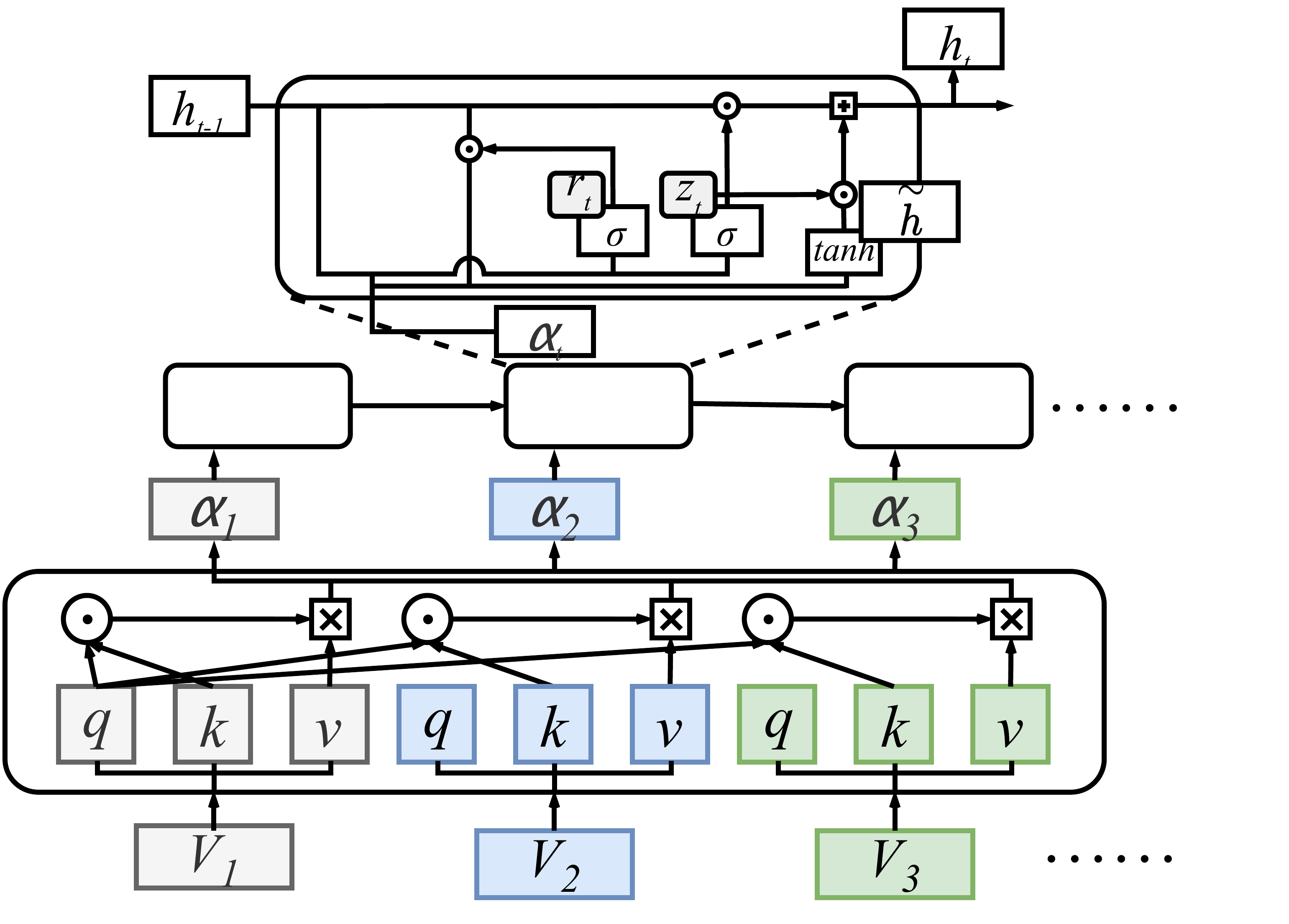}
	\caption{Structure of the implicit logic encoder, which balances the computational complexity and acquisition of higher-order information using the attention and GRU modules.}
	\label{FIG:2}
\end{figure}

Naturally, one problem is that as the number of users' historical interactions $n$ increases, the logic expressions they are converted into will increase rapidly in size. For example when $n=2$, the logic expression contains three terms (e.g., Eq. \ref{eq2}); when $n=3$, the logic expression contains seven terms (e.g., Eq. \ref{eq3}); and when $n=4$, the logic expression will contain fifteen terms. Our calculations reveal that the number of terms is $T=\sum\limits_{r=1}^{n}{\frac{n!}{r!\left( n-r \right)!}} = 2^n-1$; assuming that each term has to be disjunction calculated once, its time complexity is $O(2^n)$. This significantly reduces the computational efficiency of the model as n increases, i.e., for a large number of historical interactions. Another problem is that the DAG computation graphs constructed by the model based on the logic expressions are computed in series, which results in a constant loss of information from the forward logic units; hence, the backward logic units will inevitably have a greater impact on the final results. To overcome the two problems outlined here, we first simplify the logic expressions to prevent the computational complexity of the model from exploding. Specifically, when converting historical interactions into logic formulas, we only keep the single binary predicate terms and drop conjunction terms that consist of multiple binary predicates. We then transform the recommendation problem into the following logic query expression: 
\begin{equation}\label{eq6}
\begin{split}
    q={{v}_{?}}\cdot \exists {{u}_{i}}:Pos({{u}_{i}},{{v}_{1}})\vee Neg\left( {{u}_{i}},{{v}_{2}} \right)\vee Pos\left( {{u}_{i}},{{v}_{3}} \right) \ldots \vee Pos({{u}_{i}},{{v}_{n}})\to {{v}_{?}}
\end{split}
\end{equation}
where $n$ is the total number of historical interaction items and $T=n$ is the number of terms in the logic formula; hence, the computational complexity of the model is reduced from $O(2^n)$ to $O(n)$. Second, we construct an implicit logic encoder that captures the higher-order interactions between different logic variables using an attention mechanism to solve the information loss problem caused by ignoring the higher-order conjunction terms between each binary predicate. Moreover, the logic encoder includes a gated recurrent unit (GRU) layer, which solves the information loss problem from the serial computation disjunction paradigm by manipulating the update and reset gates. The specific structure of the implicit logic representation encoder is shown in Fig. \ref{FIG:2}.
The attention module treats all logic variables (i.e., binary predicates) in a logic expression as a sequence of inputs and captures information about the implicit interaction between different variables (i.e., those higher-order relations we discarded) by learning the correlation weights between each logic variable. This computation process mainly consists of three parts: query, key, and value. Three vectors $Q$, $K$, and $V$ are then obtained using linear projection. A similarity matrix is obtained by dot-multiplying $Q$ with $K$; the similarity formula is:
\begin{equation}\label{eq7}
\begin{split}
    f(Q,{{K}_{i}})={{Q}^{T}}{{K}_{i}}
\end{split}
\end{equation}
The weight distribution for the vector $V$ is subsequently obtained by normalizing the similarity matrix and calculating the weighted value of the vector $V$ using dot multiplication; the formula for this is given by:
\begin{equation}\label{eq8}
\begin{split}
    Attention(Q,K,V)=\text{softmax}\left( \frac{f(Q,{{K}_{i}})}{\sqrt{{{d}_{k}}}} \right)V
\end{split}
\end{equation}
To capture the complex relationships between logic variables in different subspaces, a multi-headed attention mechanism is introduced to capture different interaction information in several different projection spaces and effectively prevent the model from falling into the local optimum trap. This is represented by the following equation, where $W$ is the learnable parameter matrix.
\begin{equation}\label{eq9}
\begin{split}
hea{{d}_{i}}=Attention{{(Q,K,V)}_{i}}{{V}_{i}} \\M\!H\!ead=concat(hea{{d}_{1}},hea{{d}_{2}}\ldots hea{{d}_{i}})W 
\end{split}
\end{equation}
Using the self-attention module, we can obtain implicit logic variables ${{\alpha }_{1}}$, ${{\alpha }_{2}}$, and ${{\alpha }_{3}}$ that contain information about the higher-order interactions between the logic variables. Subsequently, a gated recurrent network is used to solve the information loss problem caused by serial computation in the deep reasoning model. The internal structure of the gated recurrent network is shown in the upper part of Fig. \ref{FIG:2}; it mainly consists of two important units: the update gate ${{z}_{t}}$ and reset gate ${{r}_{t}}$. The update gate is used to control the extent to which the information of the previous logic variable is brought into the current information, while the reset gate controls how much information from the previous state is written into the current candidate hidden state and finally outputs the weighted sequence of variables using the following calculation formula, where [·] denotes vector splicing.
\begin{equation}\label{eq10}
{{r}_{t}}=\sigma ({{W}_{r}}\cdot \text{ }\!\![\!\!\text{ }{{h}_{t-1}},{{x}_{t}}\text{ }\!\!]\!\!\text{ })
\end{equation}
\vspace*{-1em}
\begin{equation}\label{eq11}
{{z}_{t}}=\sigma ({{W}_{z}}\cdot [{{h}_{t-1}},{{x}_{t}}])
\end{equation}
\vspace*{-1em}
\begin{equation}\label{eq12}
\widetilde{h}=tanh({{W}_{\widetilde{h}}}\cdot [{{r}_{t}}*{{h}_{t-1}},{{x}_{t}}])
\end{equation}
\vspace*{-1em}
\begin{equation}\label{eq13}
{{h}_{t}}=(1-{{z}_{t}})*{{h}_{t-1}}+{{z}_{t}}*\widetilde{{{h}_{t}}}
\end{equation}
\vspace*{-1em}
\begin{equation}\label{eq14}
{{y}_{t}}=\sigma ({{W}_{0}}\cdot {{h}_{t}})
\end{equation}
Using the gated recurrent network, the model can spontaneously learn and control the trade-off of variable information at different positions of expressions, effectively improving the problem of losing variable information at the front end of logic expressions in deep inference models. Finally, a set of implicit logic variables can be output for disjunction logic computation.
\subsection{logic operation module}
In this section, we describe how the logic expressions are computed using neural networks and illustrate the training process of the model and loss function. To better understand this discussion, we strongly recommend readers to read our most relevant work NLR \cite{shi2020neural}.

\begin{figure}[ht]
	\centering
		\includegraphics[scale=.12]{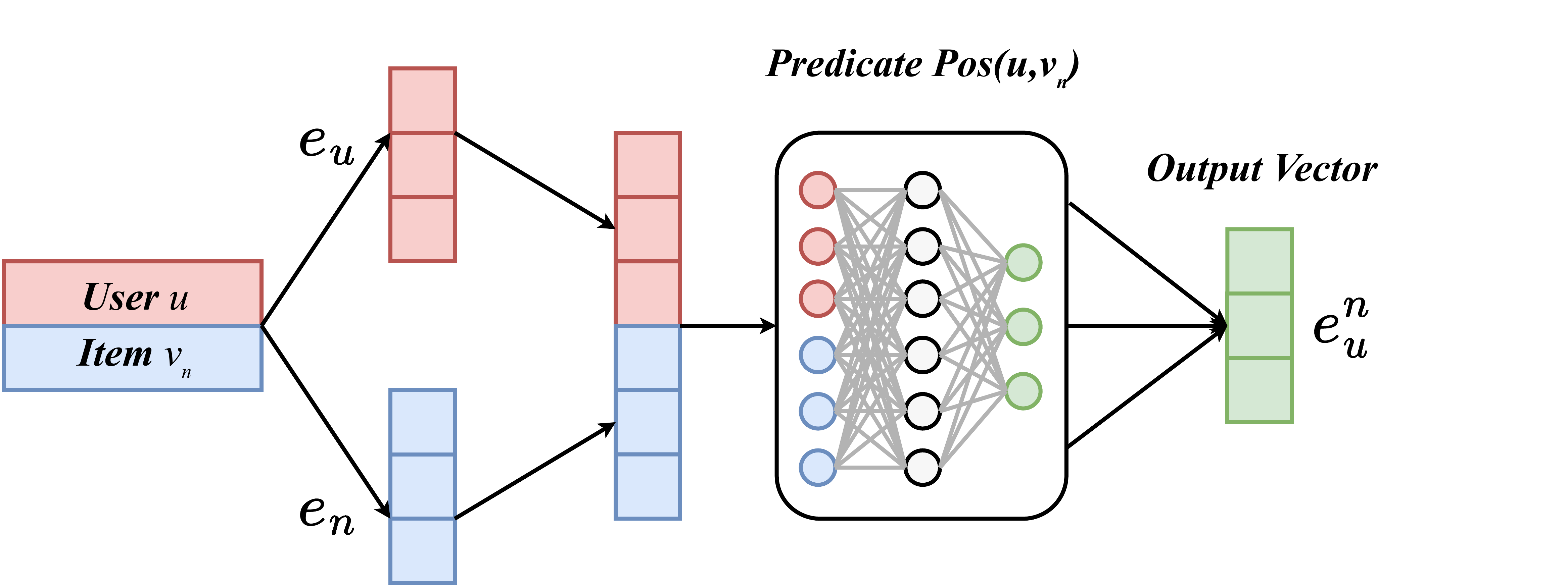}
	\caption{Modular logic operations using neural networks, where the input is a vector of two logic variables and the output is a vector of the computation result; both logic predicates and disjunction are learned using the same neural network architecture.}
	\label{FIG:3}
\end{figure}

Like most recommendation methods, we first embed users and items as vectors to perform differentiable computations. The logic operator $\vee $ and two binary predicates $Pos(.,.)$ and $Neg(.,.)$ are learned as three independent neural network modules for logic computation: $O\!R(\cdot ,\cdot )$, $P\!O\!S(\cdot ,\cdot )$, and $N\!E\!G(\cdot ,\cdot )$ for logic computation. Since Leshno et al. \cite{leshno1993multilayer} have shown that multi-layer neural networks using non-linear activation functions can theoretically approximate any function. Therefore, each operation and predicate are represented by a deep neural network: a multilayer perceptron. Consider $POS(\cdot ,\cdot )$ as an example:
\begin{equation}\label{eq15}
POS({{e}_{u}},{{e}_{n}})={{H}_{a2}}f\left( {{H}_{a1}}({{e}_{u}}\oplus {{e}_{n}})+{{b}_{a}} \right)
\end{equation}
where ${{e}_{u}}$ and ${{e}_{n}}$ represent the embedding vectors for the users and items, respectively; ${{H}_{a1}}\in {{R}^{d\times 2d}}$, ${{H}_{a2}}\in {{R}^{d\times d}}$, and ${{b}_{a}}\in {{R}^{d}}$ are parameters of the $POS(\cdot ,\cdot )$ network; $\oplus $ is the vector concat; and $f(\cdot )$ is the activation function. In this study, we used a rectified linear unit (ReLU) as the activation function. The general structure is shown in Fig. \ref{FIG:3}; the networks $NEG(\cdot ,\cdot )$ and $AND(\cdot ,\cdot )$ have the same architecture. However, their input sources are different. The implicit logic variables calculated by the implicit logic encoder are input to $AND(\cdot ,\cdot )$, and its iterative computation is based on the sequence of logic expressions; finally, the solution vector of the logic expression is output and the most similar item in the item vector space to the recommended prediction is identified. The similarity calculation formula is as follows:
\begin{equation}\label{eq16}
Sim({{e}_{q}},{{e}_{?}})=sigmoid(\varphi \frac{{{e}_{q}}\text{ }\!\!\cdot\!\!\text{ }{{e}_{?}}}{\left\| {{e}_{q}} \right\|\times \left\| {{e}_{?}} \right\|})
\end{equation}
Where ${{e}_{q}}$ is the calculated vector of logic query expressions, ${{e}_{?}}$ is the vector of candidate items, and $\varphi$ is an optional hyperparameter that enables the model to be adapted to different domain datasets. The closer the similarity is to 1, the more positive the user's expected rating for the item. The optimization goal of the model is to make the logic expression and corresponding item vectors similar to 1. We adopted a pair-wise learning strategy in training: i.e., for each item ${{v}^{+}}$ for which a user had a positive interaction with, randomly sample an item ${{v}^{-}}$ for which a user had a negative interaction with or had not yet interacted with to jointly calculate the loss function. The loss function is given by:
\begin{equation}\label{eq17}
L=-\sum\limits_{{{v}^{+}}}{\log (sigmoid(p({{v}^{+}})-p({{v}^{-}})))}
\end{equation}
where $p({{v}^{+}})$ and $p({{v}^{-}})$ are the predictions of the model. The model is encouraged to make the vector that represents a user's historical logic expressions more similar to the vector of target items than to other items to provide an accurate recommendation decision.

\subsection{Supplementary loss function}
We incorporated three additional losses to improve the performance of the model performance: rule loss, vector length loss, and parameter restriction loss. Rule loss $l_p=\sum\limits_{v\in V}{Sim(pos(u,v),neg(u,v))}$ is a modification of the predicate logic operation, where the rule $pos(u,v)=\neg neg(u,v)$ is imposed to influence the generation of the logic vector during training. Intuitively, this loss indicates that the similarity between the vectors of user-liked items and user-disliked items is expected to converge to zero; this is because we consider it impossible for a user to both like and dislike an item. The vector length loss ${l}_{\ell}=\sum\limits_{e\in E}{\left\| e \right\|_{F}^{2}}$ is a restriction on the vector length that ensures the stability of the model. The parameter restriction loss ${{l}_{\Theta}}=\left\| \Theta  \right\|_{F}^{2}$ is a restriction on all parameters in the model to prevent overfitting.

Using Eq. \ref{eq17}, the total loss function of the model is:
\begin{equation}\label{eq18}
\begin{split}
L=-\sum\limits_{{{v}^{+}}}{\log (sigmoid(p({{v}^{+}})-p({{v}^{-}})))} + {{\lambda }_{p}}\sum\limits_{v\in V}{Sim(pos(u,v),neg(u,v))}+{{\lambda }_{\ell }}\sum\limits_{e\in E}{\left\| e \right\|_{F}^{2}}+{{\lambda }_{\Theta }}\left\| \Theta  \right\|_{F}^{2} \\ 
\end{split}
\end{equation}
 where ${{\lambda }_{p}}$, ${{\lambda }_{\ell }}$, and ${{\lambda }_{\Theta }}$ are the adjustable weights of the three additional loss functions.

\section{Experiments}
In this section, we detail the extensive experiments we performed on three real-world datasets to evaluate the performance of our proposed NLQ4Rec. The experiments were conducted to answer the following five key research questions.
\begin{itemize} 
\item \textbf{RQ1}: Does NLQ4Rec yield a better recommendation performance. 
\vspace{-0.5em}
\item \textbf{RQ2}: Does the implicit logic encoder improve the ability of the model to characterize higher-order interactions? 
\vspace{-0.5em}
\item \textbf{RQ3}: Does the use of first-order logic forms and neural networks to learn binary predicate relations enhance the capability of the model? 
\vspace{-0.5em}
\item \textbf{RQ4}: Does the proposed recommendation prediction method using logic query outperform relationship-based prediction methods?
\vspace{-0.5em}
\item \textbf{RQ5}: What is the sensitivity of the hyperparameters in NLQ4Rec?
\end{itemize}
\subsection{Datasets}
We used three publicly available real-world datasets in our experiments. These three datasets cover different data volume levels and industries in which recommendation methods are used. Tab. \ref{tbl1} details some general statistical information about the three datasets.
\begin{table}[width=.9\linewidth,cols=5,pos=ht]
\caption{General statistical information about the three real-world datasets used in experiments.}\label{tbl1}
\begin{tabular*}{\tblwidth}{@{} LLLLL@{} }
\toprule
Dataset & User & Item & Interaction & Density\\
\midrule
ML100k & 943 & 1682 & 100000 & 6.30\% \\
Movies \& TV & 123961 & 50053 & 1697533 & 0.027\% \\
KindleStore & 68223 & 61934 & 982619 & 0.024\% \\
\bottomrule
\end{tabular*}
\end{table}
\begin{itemize} 
\item \textbf{ML100k} \cite{harper2015movielens}: A movie recommendation dataset was built and maintained by Grouplens, which is a common dataset in the recommendation field. It has high data density and contains 100,000 rating data published by more than 900 users for more than 1000 movies.
\vspace{-0.5em}
\item \textbf{Amazon 5-core} \cite{ni2019justifying}: A dataset from Amazon's online store that includes user, item and rating information between 1996 and 2018, which is a much larger and sparser dataset covering 29 different categories compared to ML100k. We select the \textbf{Movies \& TV} and \textbf{Kindle Store} dataset and choose each user with items that have appeared at least five times version.
\end{itemize}

Since the baseline model includes some sequential recommendation models, we included a user's earliest five interactions in the training set and their latest two positive interactions in the validation and test sets. These were preferentially assigned to the test set; no validation or test sets were generated if a user had less than five historical interactions. This is a common method for processing recommendation datasets, and is usually called the leave-one-out approach \cite{zhao2020revisiting}.
\subsection{Comparison methods}
To comprehensively evaluate the performance of our model, we compared it with nine baseline models, which can be classified into four categories: representative shallow models (BPR-MF; SVDPP), deep models (DMF; NeuMF), session-based models (STAMP; GRU4Rec; NARM), and state-of-the-art reasoning-based models (NLR; NCR). All baseline models are briefly detailed herein. 
\begin{itemize} 
\item \textbf{BPR-MF} \cite{rendle2012bpr}: A representative model for pairwise learning matrix decomposition based on Bayesian personalization ranking.
\vspace{-0.5em}
\item \textbf{SVDPP} \cite{koren2008factorization}: A combined recommendation model based on the matrix decomposition method that integrates implicit feedback and domain prediction.
\vspace{-0.5em}
\item \textbf{DMF} \cite{xue2017deep}: A deep matrix decomposition model using a deep neural network instead of a vector inner product. 
\vspace{-0.5em}
\item \textbf{NeuMF} \cite{he2017neural}: An integrated model considering features from both deep and traditional matrix decomposition, with both high- and low-dimensional features.
\vspace{-0.5em}
\item \textbf{STAMP} \cite{liu2018stamp}: A model based on short-term attention and memory priority; a popular model for obtaining users’ long- and short-term preferences using an attention mechanism.
\vspace{-0.5em}
\item \textbf{GRU4Rec} \cite{hidasi2015session}: A session-based recommendation model that uses recurrent neural networks (RNNs).
\vspace{-0.5em}
\item \textbf{NARM} \cite{li2017neural}: A model that combines an attention mechanism with a GRU and is a popular sequence recommendation model.
\vspace{-0.5em}
\item \textbf{NLR} \cite{shi2020neural}: A model based on a neural symbolic model of modular propositional logic connectives for neural networks, called Neural Logic Reasoning (NLR); this model is the most relevant to our study and is a state-of-the-art reasoning-based recommendation framework.
\vspace{-0.5em}
\item \textbf{NCR} \cite{chen2021neural}: A personalized neural symbolic recommendation model based on NLR that additionally considers user information, called Neural Collaborative Reasoning (NCR); this model is a state-of-the-art reasoning-based recommendation framework.
\end{itemize}
\subsection{Parameter settings}
All models were trained with 500 epochs using the Adam optimizer and a batch-size of 128. The learning rate was 0.0001 and early-stopping was conducted according to the performance on the validation set. Both ${{\lambda }_{\ell }}$ and ${{\lambda }_{\Theta }}$ were set to $1\times10^{-4}$ and applied to the baseline models equally; ${{\lambda }_{p}}$ was set to $1\times10^{-5}$. The vector embedding dimension was set to 64 for all baseline models except DMF, for which it was set to 128. The results in Tab. \ref{tbl2} report the performance when the vector embedding dimension was set to 96 for the proposed model; this is the dimension we set when we first showed progress in our exploration, although it is not the best embedding dimension for our method. Performance data of our model for different embedding dimensions is also provided herein (see the detailed report in Section 4.7). The maximum history interaction length was set to 10 for session-based and reasoning-based models (including the proposed NLQ4Rec). Three different random seeds were used in the experiments; and the average results for each model are reported. Source code is available on GitHub, more details of the parameter settings can be found on there\footnote{https://github.com/hanzo2020/NLQ}.
\subsection{Evaluation metric}
Following previous works \cite{zhao2020revisiting}, we adopted two widely used metrics to measure the recommendation performance of all the models: Normalized Discounted Cumulative Gain (NDCG) and hit ratio (HR). In the validation and test experiments, for each item that users gave a positive rating, 100 items that users gave negative ratings or had not interacted with were randomly selected to generate a test sequence; these 101 items were then ranked for recommendation.
\begin{itemize} 
\item \textbf{NDCG@K}: A larger result indicates a higher position of the target item in the sequence of the top K items; the calculation is given by: 
\begin{equation}\label{eq19}
N\!DCG@K=\frac{1}{N}\sum\limits_{N}{\frac{{{\log }_{e}}2}{{{\log }_{e}}(i+2)}}
\end{equation}
\vspace{-0.5em}
\item \textbf{HR@K}: A hit indicates whether a target item appears in the top K items; 1 indicates occurrence, while 0 is used otherwise. HR@K represents the probability of successfully predicting the target item in all test sets; the calculation is given by:
\begin{equation}\label{eq20}
H\!R@K=\frac{number\text{ }of\text{ }hits}{N}
\end{equation}
\end{itemize}
\subsection{Recommendation performance(RQ1)}
In this section, we present the experimental results and analyze them using NDCG@K and HIT@K. In Tab. \ref{tbl2}, the best results from the matching-based baselines (i.e., the first seven baselines) in each column are underlined and the best results from all baseline models (including the two reasoning-based baseline models: NLR and NCR) are underlined with a wavy line. The best results for each entire column are highlighted in bold.
\begin{table}[width=.9\linewidth,cols=13,pos=ht]
\captionsetup{justification=raggedright}
\caption{Recommendation performance (NDCG and HR) of the proposed and baseline models on three datasets. The best results from the matching-based baselines (i.e., the first seven baselines) in each column are \underline{underlined} and the best results from all baseline models (including the two reasoning-based baseline models: NLR and NCR) are underlined with a \uwave{wavy line}. The best results for each entire column are highlighted in bold.  Improve1 and Improve2 are the percentage improvement of the results of the proposed model compared with the best results of the matching-based and reasoning-based baseline models, respectively.}\label{tbl2}
\setlength{\tabcolsep}{4mm}
\resizebox{\textwidth}{!}{
\large
\begin{tabular}{lllllllllllll}
\toprule
 & \multicolumn{4}{c}{\pmb{ML100k}} & \multicolumn{4}{c}{\pmb{Movies and TV}} & \multicolumn{4}{c}{\pmb{KindleStore}}\\
\cmidrule(r){2-5}\cmidrule(r){6-9}\cmidrule(r){10-13}
 & N@5 & N@10 & HR@5 & HR@10 & N@5 & N@10 & HR@5 & HR@10 & N@5 & N@10 & HR@5 & HR@10 \\
\midrule
BPRMF & 0.3051 & 0.3599 & 0.4566 & 0.6265 & 0.4037 & 0.4460 & 0.5421 & 0.6728 & 0.4626 & 0.5024 & 0.6120 & 0.7343\\
SVDPP & 0.3168 & 0.3778 & 0.4732 & 0.6624 & 0.4157 & 0.4569 & 0.5477 & 0.6770 & \uwave{\underline{0.4815}}\par & \uwave{\underline{0.5205}}\par & \underline{0.6217} & \underline{0.7422}\\
\midrule
DMF & 0.3023 & 0.3643 & 0.4489 & 0.6488 & 0.4059 & 0.4445 & 0.5469 & 0.6775 & 0.4489 & 0.4975 & 0.5972 & 0.7194\\
NCF & 0.3041 & 0.3595 & 0.4512 & 0.6338 & 0.3943 & 0.4398 & 0.5242 & 0.6645 & 0.4568 & 0.5059 & 0.6038 & 0.7259\\
\midrule
STAMP & 0.3353 & 0.3907 & 0.4893 & 0.6602 & 0.4047 & 0.4458 & 0.5385 & 0.6655 & 0.4517 & 0.4906 & 0.5923 & 0.7121\\
GRU4Rec & \uwave{\underline{0.3536}}\par & \uwave{\underline{0.4094}}\par & \underline{0.5129} & \underline{0.6838} & 0.4209 & \underline{0.4616} & \underline{0.5560} & \underline{0.6818} & 0.4782 & 0.5138 & 0.6159 & 0.7253\\
NARM & 0.3511 & 0.4084 & 0.5065 & 0.6833 & \underline{0.4210} & 0.4612 & 0.5533 & 0.6797 & 0.4619 & 0.4982 & 0.5984 & 0.7101\\
\midrule
NLR & 0.3466 & 0.4059 & 0.4995 & 0.6672 & 0.4220 & 0.4626 & 0.5567 & 0.6822 & 0.4539 & 0.4934 & 0.6066 & 0.7225\\
NCR & 0.3504 & 0.4064 & \uwave{0.5133}\par & \uwave{0.6948}\par & \uwave{0.4374}\par & \uwave{0.4767}\par & \uwave{0.5728}\par & \uwave{0.6964}\par & 0.4708 & 0.5139 & \uwave{0.6218}\par & \uwave{0.7426}\par\\
\midrule
NLQ4Rec & \pmb{0.3792} & \pmb{0.4316} & \pmb{0.5504} & \pmb{0.7128} & \pmb{0.4654} & \pmb{0.5033} & \pmb{0.5997} & \pmb{0.7168} & \pmb{0.5538} & \pmb{0.5864} & \pmb{0.6908} & \pmb{0.7909}\\
\midrule
\midrule
improve1 & 7.23\% & 5.42\% & 7.31\% & 4.23\% & 10.55\% & 9.03\% & 7.87\% & 5.13\% & 15.01\% & 12.66\% & 11.12\% & 6.56\%\\
improve2 & 8.20\% & 6.19\% & 7.22\% & 2.58\% & 6.40\% & 5.58\% & 4.71\% & 2.93\% & 17.63\% & 14.11\% & 11.09\% & 6.50\%\\
\bottomrule
\end{tabular}
}
\end{table}

Our approach achieves the best performance on all three datasets, and the best reasoning-based approach outperforms the matching-based approaches on most metrics. GRU4Rec outperforms the other matching-based methods on the ML100k and Movies \& TV datasets; NARM yields a similar performance. These models also exploit temporal information in the recommendation data, unlike the otherwise-similar session-based recommendation model STAMP; the comparison shows that temporal features can significantly improve recommendation performance. Although the matrix decomposition-based model performed poorly on the ML100k and Movies \& TV datasets, it narrowed the gap with other models on the KindleStore dataset. The SVDPP model yielded the best performance of the matching-based baselines on this dataset. The reason for this is that the total number of users and items in this dataset is less disparate, and the shape of the composed interaction matrix is closer to that of a square matrix, which may be more favorable for matrix decomposition methods. Compared to other matrix decomposition methods, SVDPP can utilize both display and implicit feedback information and therefore performs better on this dataset. Compared to these baselines, the cognitive reasoning ability in the proposed model yields a significant improvement in performance; improve1 in Tab. \ref{tbl2} shows the percentage improvement of the proposed model compared with the best matching-based baseline model.

We also incorporated reasoning-based baselines for comparison. Both NLR and NCR show a good performance. NCR is superior to the NLR and yields the best performance of all baseline models in most metrics, which is due to the ability of the NCR to incorporate logic-based reasoning in its predictions, as it additionally considers the users during event embedding for personalized learning. However, the proposed model adds an implicit logic encoder to balance computational complexity with feature extraction capability and further refines the representation of user history data by introducing predicate logic and reasoning in the form of queries based on logic expressions to make the model more suitable for recommendation prediction tasks. In Tab. \ref{tbl2}, improve2 shows the percentage improvement of the proposed model compared with the best result from the reasoning-based models, which is the result obtained by the NCR model.

\subsection{Ablation experiment(RQ2-4)}
To further compare the proposed model with each baseline model, we conducted ablation experiments for each of the three improvements proposed in this paper. The results are shown in Tab. \ref{tbl3}. NLQ4Rec-q, -e, and -p correspond to the proposed model without the logic query, implicit logic encoders, and predicate modules, respectively. When the logic query is removed, the true and false values of the logic expressions are calculated as in NLR. When the implicit logic encoder is removed, the reduced complexity logic expressions are used for calculations. When the predicate module is removed, only commodity sequences are considered and the NOT network from the NLR model is used to represent negative interactions.
\begin{table}[width=.9\linewidth,cols=13,pos=ht]
\caption{Ablation comparisons of NLR4Rec: model performance without the logic query (-q), implicit logic encoders (-e), and predicate network computations (-p). Improve-q, -e, and -p indicate the percentage improvement of the full proposed model compared with each corresponding reduced model.}\label{tbl3}
\setlength{\tabcolsep}{4mm}
\resizebox{\textwidth}{!}{
\large
\begin{tabular}{lllllllllllll}
\toprule
 & \multicolumn{4}{c}{\pmb{ML100k}} & \multicolumn{4}{c}{\pmb{Movies and TV}} & \multicolumn{4}{c}{\pmb{KindleStore}}\\
\cmidrule(r){2-5}\cmidrule(r){6-9}\cmidrule(r){10-13}
 & N@5 & N@10 & HR@5 & HR@10 & N@5 & N@10 & HR@5 & HR@10 & N@5 & N@10 & HR@5 & HR@10 \\
\midrule
NLQ4Rec-q & 0.3555 & 0.4075 & 0.5166 & 0.6753 & 0.4392 & 0.4788 & 0.5743 & 0.6969 & 0.4897 & 0.5289 & 0.6365 & 0.7568\\
NLQ4Rec-e & 0.3645 & 0.4174 & 0.5263 & 0.6898 & 0.4568 & 0.4951 & 0.5893 & 0.7074 & 0.5339 & 0.5672 & 0.6712 & 0.7735\\
NLQ4Rec-p & 0.3677 & 0.4174 & 0.5359 & 0.6887 & 0.4482 & 0.4872 & 0.5817 & 0.7021 & 0.5301 & 0.5636 & 0.6657 & 0.7688\\
NLQ4Rec & \pmb{0.3792} & \pmb{0.4316} & \pmb{0.5504} & \pmb{0.7128} & \pmb{0.4654} & \pmb{0.5033} & \pmb{0.5997} & \pmb{0.7168} & \pmb{0.5538} & \pmb{0.5864} & \pmb{0.6908} & \pmb{0.7909}\\
\midrule
\midrule
improve-q & 6.67\% & 5.91\% & 6.53\% & 5.55\% & 5.98\% & 5.12\% & 4.42\% & 2.85\% & 13.08\% & 10.87\% & 8.53\% & 4.50\%\\
improve-e & 4.02\% & 3.40\% & 4.58\% & 3.33\% & 1.89\% & 1.67\% & 1.76\% & 1.33\% & 3.72\% & 3.38\% & 2.92\% & 2.24\%\\
improve-p & 3.11\% & 3.48\% & 2.70\% & 3.50\% & 3.84\% & 3.30\% & 3.09\% & 2.09\% & 4.47\% & 4.04\% & 3.76\% & 2.87\%\\
\bottomrule
\end{tabular}
}
\end{table}

In Tab. \ref{tbl3}, improve-q, -e, and -p represent the percentage improvement of the full proposed model compared with each corresponding reduced model. These values could be considered to be the percentage contribution of each part to the overall improvement in performance of the model. All three of our proposed innovations are validated; the logic query has the highest impact on the performance of the model while the implicit logic encoder has the lowest. The query-based approach is clearly more suitable for the decision reasoning behavior in the recommendation and improves the precision, which is why Neural Logic Query (NLQ) was used in the name of the model.
\subsection{Parameter sensitivity analysis(RQ5)}
\begin{figure}[ht]
	\centering
		\includegraphics[scale=.35]{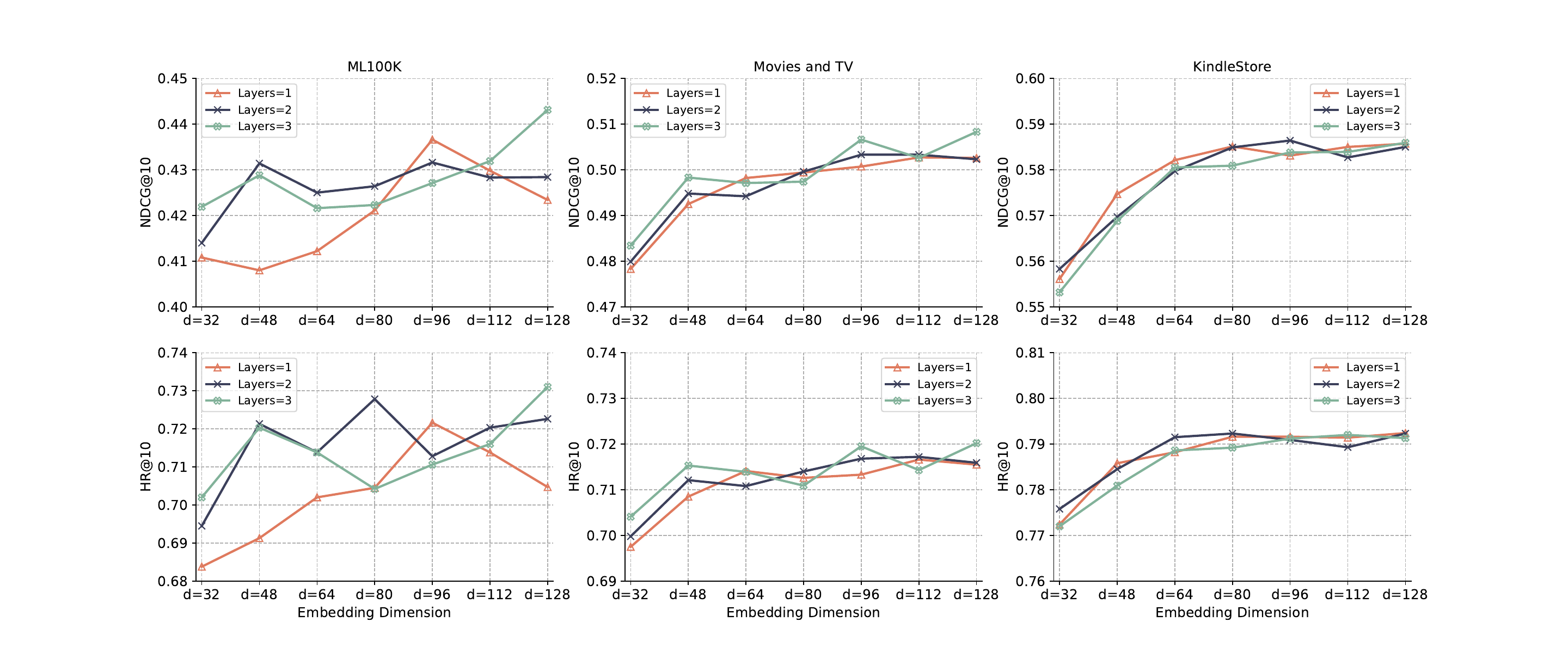}
	\caption{NLR4Rec performance for different vector dimensions (d) and different numbers of self-attention layers (l).}
	\label{FIG:4}
\end{figure}
\begin{figure}[ht]
	\centering
		\includegraphics[scale=.35]{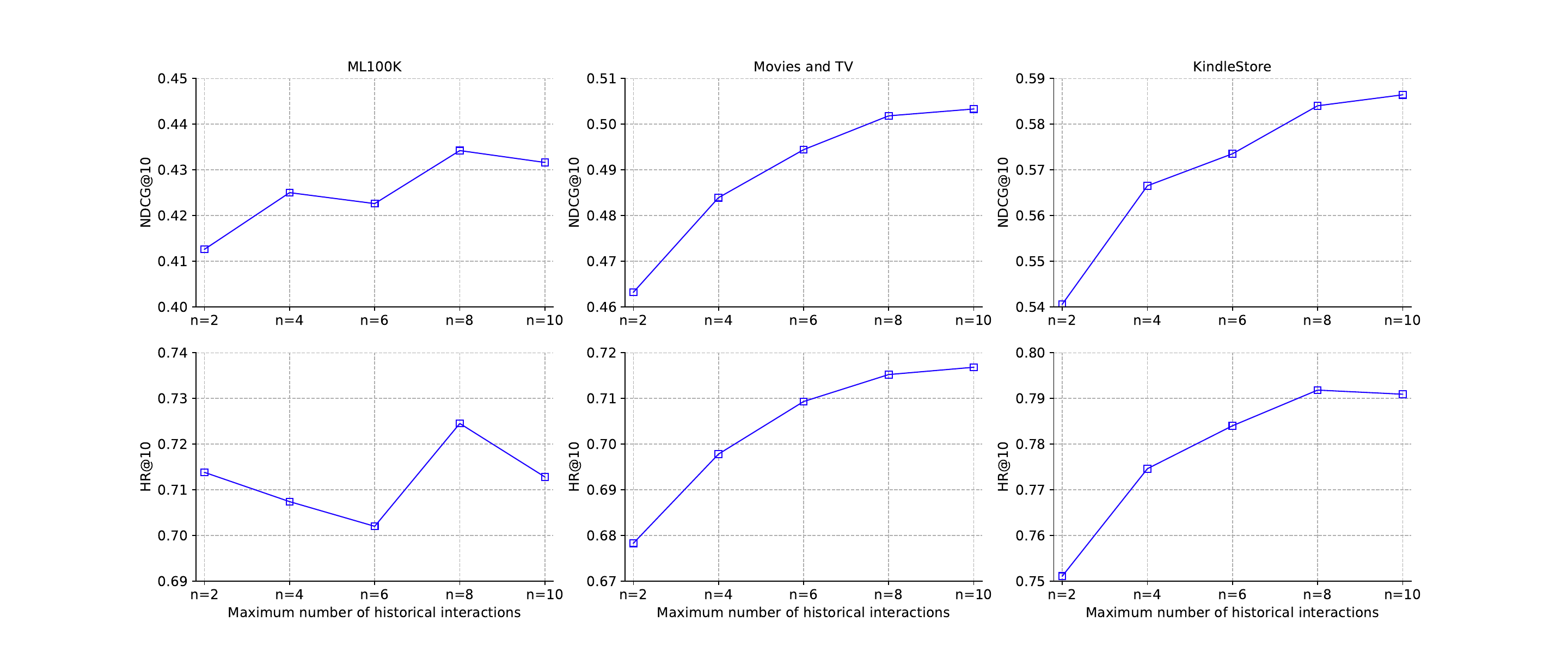}
	\caption{NLR4Rec performance for different values of the maximum number of historical interactions (n).}
	\label{FIG:5}
\end{figure}
We also performed parameter sensitivity experiments for the following parameters: the embedding dimension of the model (d), number of layers of the self-attention module in the implicit logic encoder (l), maximum number of historical interactions (n), and logic rule weight ${{\lambda }_{p}}$. 

The embedding dimension of the model was varied from 32 to128 and the number of layers of the self-attention module was varied from 1 to 3; the experimental results are shown in Fig. \ref{FIG:4}. The performance of the model decreases significantly when the dimension is too low, especially for only one attention layer. The model achieves optimal performance for an embedding dimension of 128 with three attention layers. This indicates that the capturing of higher-order interactions between variables by the implicit logic encoder is influenced by the representation capacity and number of attention layers. In general, the performance of the model stabilizes when the number of attention layers is greater than 1 and the dimension is greater than or equal to 64. The performance of the model may improve with more attention layers or larger dimensions, however this will significantly increase the computation time and test the GPU memory; readers can use the code provided for further experimentation if they are interested in exploring this.
\begin{figure}[ht]
	\centering
		\includegraphics[scale=.35]{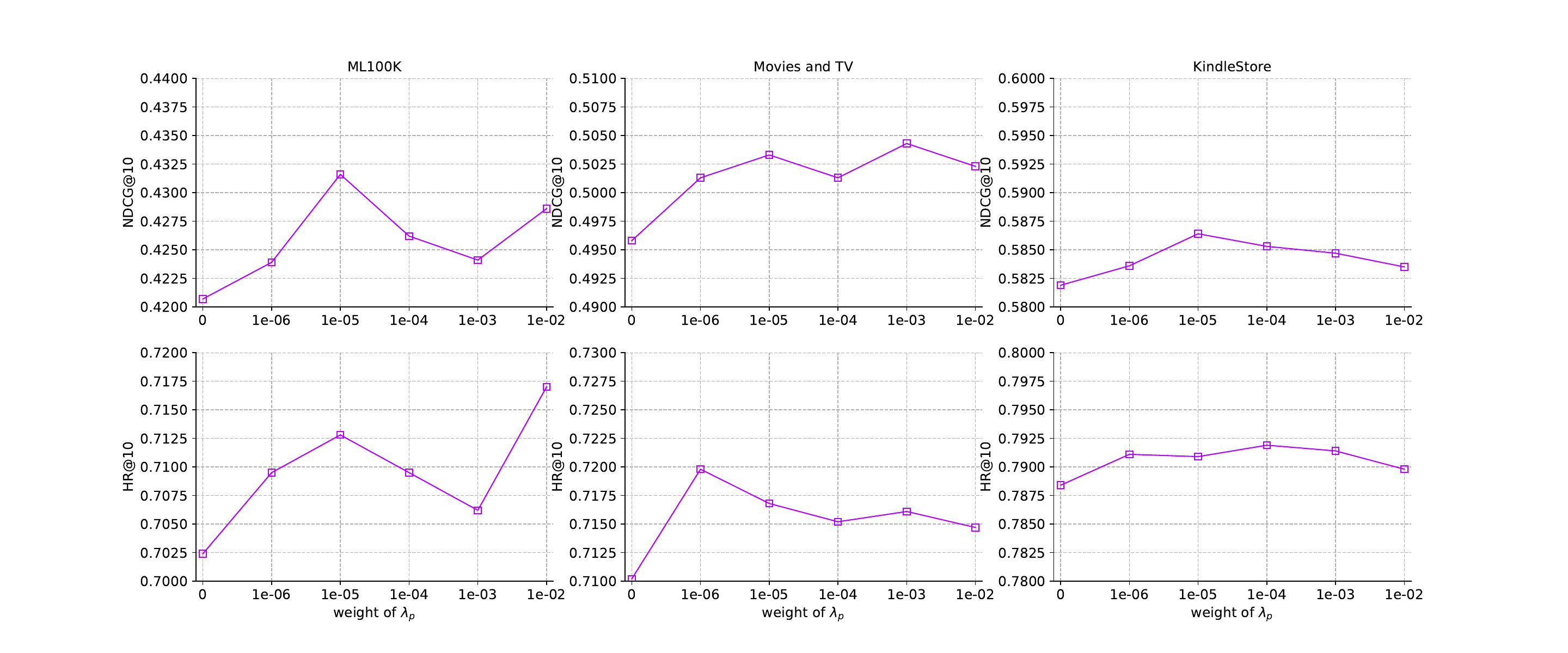}
	\caption{NLR4Rec performance for different rule weights ${{\lambda }_{p}}$.}
	\label{FIG:6}
\end{figure}

The maximum number of historical interactions n was varied from 2 to 10; the experimental results are shown in Fig. \ref{FIG:5}. The performance of the model improves as n increases. This is because more historical interactions provide more logic reasoning rules for the model to learn. However, only 8 historical interactions are needed to achieve the best performance on some datasets; in these cases, the performance decreases when n increases to 10. A possible reason for this is that too many historical interactions may result in more internal conflicts or noise, especially when many users do not have enough historical interactions. In this case, incorporating more and more complex logic rules into the training set will have a negative effect, e.g., overfitting. The potential complexity of this mechanism may need to be explored in the future.

The rule loss weight ${{\lambda }_{p}}$ was varied from 0 to 0.01 to investigate our theory that the reasoning performance of the model will improve when the rule $pos(u,v)=\neg neg(u,v)$ is used to influence the generation of the logic vector during training. The results of the experiments for different loss weights are shown in Fig. \ref{FIG:6}. The performance of the model with a loss weight of 0 was lower than that for other values, indicating that this rule does improve the performance of the model. However, a high weight does not yield further improvement in the accuracy of the recommendation; this is potentially because recommendation is not a purely logic reasoning task: a user who does not like Iron Man and another user who likes Iron Man may both like Transformers. Our experiments show that the model performs the best when the loss weight is set to 1e-05.

\section{Conclusion}
We proposed a reasoning-based neuro-symbolic recommendation method, called neural logic query for recommendation (NLQ4Rec), to address the lack of cognitive reasoning capabilities in current recommendation models. We adopted a decision perspective for recommendation prediction; the method first transforms a user's historical interactions (i.e., historical behaviors) into a set of logic rules and then treats the recommendation task as a logic-based query problem. Then, modular logic operations based on neural networks and queries are performed in the vector space to predict the user's future items of interest (i.e., the user's future decisions). We also optimized the neural computation for logic expressions by constructing implicit logic encoders. Experiments on three real-world datasets demonstrated that our model significantly outperforms seven matching-based recommendation models and two state-of-the-art reasoning-based models. We also performed ablation experiments to further validate the effectiveness of the main concepts in the proposed model.

In the future, we will continue to improve the model in three ways. First, we will consider how to incorporate more prior information from external sources, such as contextual and multimodal information for reasoning. Second, we aim to provide some principle explainability to the model, to automatically generate explainable computational guidelines on non-predefined templates. Finally, we hope to explore potential applications of the model in areas other than recommendation, especially for tasks that predict future actions based on past behaviors (e.g., decision problems). An interesting example is the game of Go, where each available point for a move is a potential recommendation item, and the best move must be recommended based on the historical behavior of two players. Knowledge graph reasoning tasks combined with medicine and law also offer very promising application scenarios.
\printcredits
\section*{Declaration of competing interest}
The authors declare that they have no known competing financial interests or personal relationships that could have appeared to influence the work reported in this paper.
\section*{Acknowledgements}
This work was supported by the National Natural Science Foundation of China (No. 61906066), Natural Science Foundation of Zhejiang Province (No. LQ18F020002), Zhejiang Provincial Education Department Scientific Research Project(No. Y202044192), Postgraduate Research and Innovation Project of Huzhou University (No. 2022KYCX43).
\bibliographystyle{model1-num-names}

\bibliography{cas-sc-template}

\begin{thebibliography}{59}
\expandafter\ifx\csname natexlab\endcsname\relax\def\natexlab#1{#1}\fi
\providecommand{\url}[1]{\texttt{#1}}
\providecommand{\href}[2]{#2}
\providecommand{\path}[1]{#1}
\providecommand{\DOIprefix}{doi:}
\providecommand{\ArXivprefix}{arXiv:}
\providecommand{\URLprefix}{URL: }
\providecommand{\Pubmedprefix}{pmid:}
\providecommand{\doi}[1]{\href{http://dx.doi.org/#1}{\path{#1}}}
\providecommand{\Pubmed}[1]{\href{pmid:#1}{\path{#1}}}
\providecommand{\bibinfo}[2]{#2}
\ifx\xfnm\relax \def\xfnm[#1]{\unskip,\space#1}\fi
\bibitem[{Cui et~al.(2020)Cui, Xu, Fei, Cai, Cao, Zhang, and Chen}]{cui2020personalized}
\bibinfo{author}{Z.~Cui}, \bibinfo{author}{X.~Xu}, \bibinfo{author}{X.~Fei}, \bibinfo{author}{X.~Cai}, \bibinfo{author}{Y.~Cao}, \bibinfo{author}{W.~Zhang}, \bibinfo{author}{J.~Chen},
\newblock \bibinfo{title}{Personalized recommendation system based on collaborative filtering for iot scenarios},
\newblock \bibinfo{journal}{IEEE Transactions on Services Computing} \bibinfo{volume}{13} (\bibinfo{year}{2020}) \bibinfo{pages}{685--695}.
\bibitem[{Schafer et~al.(2007)Schafer, Frankowski, Herlocker, and Sen}]{schafer2007collaborative}
\bibinfo{author}{J.~B. Schafer}, \bibinfo{author}{D.~Frankowski}, \bibinfo{author}{J.~Herlocker}, \bibinfo{author}{S.~Sen},
\newblock \bibinfo{title}{Collaborative filtering recommender systems},
\newblock in: \bibinfo{booktitle}{The adaptive web}, \bibinfo{publisher}{Springer}, \bibinfo{year}{2007}, pp. \bibinfo{pages}{291--324}.
\bibitem[{Sarwar et~al.(2001)Sarwar, Karypis, Konstan, and Riedl}]{sarwar2001item}
\bibinfo{author}{B.~Sarwar}, \bibinfo{author}{G.~Karypis}, \bibinfo{author}{J.~Konstan}, \bibinfo{author}{J.~Riedl},
\newblock \bibinfo{title}{Item-based collaborative filtering recommendation algorithms},
\newblock in: \bibinfo{booktitle}{Proceedings of the 10th international conference on World Wide Web}, \bibinfo{year}{2001}, pp. \bibinfo{pages}{285--295}.
\bibitem[{Resnick et~al.(1994)Resnick, Iacovou, Suchak, Bergstrom, and Riedl}]{resnick1994grouplens}
\bibinfo{author}{P.~Resnick}, \bibinfo{author}{N.~Iacovou}, \bibinfo{author}{M.~Suchak}, \bibinfo{author}{P.~Bergstrom}, \bibinfo{author}{J.~Riedl},
\newblock \bibinfo{title}{Grouplens: An open architecture for collaborative filtering of netnews},
\newblock in: \bibinfo{booktitle}{Proceedings of the 1994 ACM conference on Computer supported cooperative work}, \bibinfo{year}{1994}, pp. \bibinfo{pages}{175--186}.
\bibitem[{Koren et~al.(2009)Koren, Bell, and Volinsky}]{koren2009matrix}
\bibinfo{author}{Y.~Koren}, \bibinfo{author}{R.~Bell}, \bibinfo{author}{C.~Volinsky},
\newblock \bibinfo{title}{Matrix factorization techniques for recommender systems},
\newblock \bibinfo{journal}{Computer} \bibinfo{volume}{42} (\bibinfo{year}{2009}) \bibinfo{pages}{30--37}.
\bibitem[{He et~al.(2016)He, Zhang, Kan, and Chua}]{he2016fast}
\bibinfo{author}{X.~He}, \bibinfo{author}{H.~Zhang}, \bibinfo{author}{M.-Y. Kan}, \bibinfo{author}{T.-S. Chua},
\newblock \bibinfo{title}{Fast matrix factorization for online recommendation with implicit feedback},
\newblock in: \bibinfo{booktitle}{Proceedings of the 39th International ACM SIGIR conference on Research and Development in Information Retrieval}, \bibinfo{year}{2016}, pp. \bibinfo{pages}{549--558}.
\bibitem[{Tay et~al.(2018)Tay, Anh~Tuan, and Hui}]{tay2018latent}
\bibinfo{author}{Y.~Tay}, \bibinfo{author}{L.~Anh~Tuan}, \bibinfo{author}{S.~C. Hui},
\newblock \bibinfo{title}{Latent relational metric learning via memory-based attention for collaborative ranking},
\newblock in: \bibinfo{booktitle}{Proceedings of the 2018 world wide web conference}, \bibinfo{year}{2018}, pp. \bibinfo{pages}{729--739}.
\bibitem[{Yin et~al.(2019)Yin, Li, Zhang, and Lu}]{yin2019deeper}
\bibinfo{author}{R.~Yin}, \bibinfo{author}{K.~Li}, \bibinfo{author}{G.~Zhang}, \bibinfo{author}{J.~Lu},
\newblock \bibinfo{title}{A deeper graph neural network for recommender systems},
\newblock \bibinfo{journal}{Knowledge-Based Systems} \bibinfo{volume}{185} (\bibinfo{year}{2019}) \bibinfo{pages}{105020}.
\bibitem[{Xue et~al.(2017)Xue, Dai, Zhang, Huang, and Chen}]{xue2017deep}
\bibinfo{author}{H.-J. Xue}, \bibinfo{author}{X.~Dai}, \bibinfo{author}{J.~Zhang}, \bibinfo{author}{S.~Huang}, \bibinfo{author}{J.~Chen},
\newblock \bibinfo{title}{Deep matrix factorization models for recommender systems.},
\newblock in: \bibinfo{booktitle}{IJCAI}, volume~\bibinfo{volume}{17}, \bibinfo{organization}{Melbourne, Australia}, \bibinfo{year}{2017}, pp. \bibinfo{pages}{3203--3209}.
\bibitem[{Tian et~al.(2022)Tian, Guo, Li, Liu, and Wang}]{tian2022exploiting}
\bibinfo{author}{S.~Tian}, \bibinfo{author}{G.~Guo}, \bibinfo{author}{Y.~Li}, \bibinfo{author}{Y.~Liu}, \bibinfo{author}{X.~Wang},
\newblock \bibinfo{title}{Exploiting high-order local and global user--item interactions for effective recommendation},
\newblock \bibinfo{journal}{Knowledge-Based Systems} \bibinfo{volume}{246} (\bibinfo{year}{2022}) \bibinfo{pages}{108618}.
\bibitem[{Li et~al.(2017)Li, Ren, Chen, Ren, Lian, and Ma}]{li2017neural}
\bibinfo{author}{J.~Li}, \bibinfo{author}{P.~Ren}, \bibinfo{author}{Z.~Chen}, \bibinfo{author}{Z.~Ren}, \bibinfo{author}{T.~Lian}, \bibinfo{author}{J.~Ma},
\newblock \bibinfo{title}{Neural attentive session-based recommendation},
\newblock in: \bibinfo{booktitle}{Proceedings of the 2017 ACM on Conference on Information and Knowledge Management}, \bibinfo{year}{2017}, pp. \bibinfo{pages}{1419--1428}.
\bibitem[{Heidari et~al.(2022)Heidari, Moradi, and Koochari}]{heidari2022attention}
\bibinfo{author}{N.~Heidari}, \bibinfo{author}{P.~Moradi}, \bibinfo{author}{A.~Koochari},
\newblock \bibinfo{title}{An attention-based deep learning method for solving the cold-start and sparsity issues of recommender systems},
\newblock \bibinfo{journal}{Knowledge-Based Systems} \bibinfo{volume}{256} (\bibinfo{year}{2022}) \bibinfo{pages}{109835}.
\bibitem[{Tang et~al.(2022)Tang, Zhu, Guo, and Dietze}]{tang2022time}
\bibinfo{author}{G.~Tang}, \bibinfo{author}{X.~Zhu}, \bibinfo{author}{J.~Guo}, \bibinfo{author}{S.~Dietze},
\newblock \bibinfo{title}{Time enhanced graph neural networks for session-based recommendation},
\newblock \bibinfo{journal}{Knowledge-Based Systems}  (\bibinfo{year}{2022}) \bibinfo{pages}{109204}.
\bibitem[{Zhao et~al.(2020)Zhao, Lou, Qian, and Hou}]{zhao2020personalized}
\bibinfo{author}{G.~Zhao}, \bibinfo{author}{P.~Lou}, \bibinfo{author}{X.~Qian}, \bibinfo{author}{X.~Hou},
\newblock \bibinfo{title}{Personalized location recommendation by fusing sentimental and spatial context},
\newblock \bibinfo{journal}{Knowledge-Based Systems} \bibinfo{volume}{196} (\bibinfo{year}{2020}) \bibinfo{pages}{105849}.
\bibitem[{Zhang et~al.(2016)Zhang, Yuan, Lian, Xie, and Ma}]{zhang2016collaborative}
\bibinfo{author}{F.~Zhang}, \bibinfo{author}{N.~J. Yuan}, \bibinfo{author}{D.~Lian}, \bibinfo{author}{X.~Xie}, \bibinfo{author}{W.-Y. Ma},
\newblock \bibinfo{title}{Collaborative knowledge base embedding for recommender systems},
\newblock in: \bibinfo{booktitle}{Proceedings of the 22nd ACM SIGKDD international conference on knowledge discovery and data mining}, \bibinfo{year}{2016}, pp. \bibinfo{pages}{353--362}.
\bibitem[{Wang et~al.(2022)Wang, Xu, Li, Li, and Wang}]{wang2022tkgat}
\bibinfo{author}{B.~Wang}, \bibinfo{author}{H.~Xu}, \bibinfo{author}{C.~Li}, \bibinfo{author}{Y.~Li}, \bibinfo{author}{M.~Wang},
\newblock \bibinfo{title}{Tkgat: Graph attention network for knowledge-enhanced tag-aware recommendation system},
\newblock \bibinfo{journal}{Knowledge-Based Systems} \bibinfo{volume}{257} (\bibinfo{year}{2022}) \bibinfo{pages}{109903}.
\bibitem[{Cheng et~al.(2016)Cheng, Koc, Harmsen, Shaked, Chandra, Aradhye, Anderson, Corrado, Chai, Ispir et~al.}]{cheng2016wide}
\bibinfo{author}{H.-T. Cheng}, \bibinfo{author}{L.~Koc}, \bibinfo{author}{J.~Harmsen}, \bibinfo{author}{T.~Shaked}, \bibinfo{author}{T.~Chandra}, \bibinfo{author}{H.~Aradhye}, \bibinfo{author}{G.~Anderson}, \bibinfo{author}{G.~Corrado}, \bibinfo{author}{W.~Chai}, \bibinfo{author}{M.~Ispir}, et~al.,
\newblock \bibinfo{title}{Wide \& deep learning for recommender systems},
\newblock in: \bibinfo{booktitle}{Proceedings of the 1st workshop on deep learning for recommender systems}, \bibinfo{year}{2016}, pp. \bibinfo{pages}{7--10}.
\bibitem[{Marcus(2020)}]{marcus2020next}
\bibinfo{author}{G.~Marcus},
\newblock \bibinfo{title}{The next decade in ai: four steps towards robust artificial intelligence},
\newblock \bibinfo{journal}{arXiv preprint arXiv:2002.06177}  (\bibinfo{year}{2020}).
\bibitem[{Chen et~al.(2021)Chen, Shi, Li, and Zhang}]{chen2021neural}
\bibinfo{author}{H.~Chen}, \bibinfo{author}{S.~Shi}, \bibinfo{author}{Y.~Li}, \bibinfo{author}{Y.~Zhang},
\newblock \bibinfo{title}{Neural collaborative reasoning},
\newblock in: \bibinfo{booktitle}{Proceedings of the Web Conference 2021}, \bibinfo{year}{2021}, pp. \bibinfo{pages}{1516--1527}.
\bibitem[{Ferrari~Dacrema et~al.(2019)Ferrari~Dacrema, Cremonesi, and Jannach}]{ferrari2019we}
\bibinfo{author}{M.~Ferrari~Dacrema}, \bibinfo{author}{P.~Cremonesi}, \bibinfo{author}{D.~Jannach},
\newblock \bibinfo{title}{Are we really making much progress? a worrying analysis of recent neural recommendation approaches},
\newblock in: \bibinfo{booktitle}{Proceedings of the 13th ACM conference on recommender systems}, \bibinfo{year}{2019}, pp. \bibinfo{pages}{101--109}.
\bibitem[{Rendle et~al.(2020)Rendle, Krichene, Zhang, and Anderson}]{rendle2020neural}
\bibinfo{author}{S.~Rendle}, \bibinfo{author}{W.~Krichene}, \bibinfo{author}{L.~Zhang}, \bibinfo{author}{J.~Anderson},
\newblock \bibinfo{title}{Neural collaborative filtering vs. matrix factorization revisited},
\newblock in: \bibinfo{booktitle}{Fourteenth ACM conference on recommender systems}, \bibinfo{year}{2020}, pp. \bibinfo{pages}{240--248}.
\bibitem[{Bengio(2019)}]{bengio2019system}
\bibinfo{author}{Y.~Bengio},
\newblock \bibinfo{title}{From system 1 deep learning to system 2 deep learning},
\newblock in: \bibinfo{booktitle}{Neural Information Processing Systems}, \bibinfo{year}{2019}.
\bibitem[{Harmelen(2022)}]{harmelen2022preface}
\bibinfo{author}{F.~Harmelen},
\newblock \bibinfo{title}{Preface: The 3rd ai wave is coming, and it needs a theory},
\newblock in: \bibinfo{booktitle}{Neuro-Symbolic Artificial Intelligence: The State of the Art}, \bibinfo{publisher}{IOS Press BV}, \bibinfo{year}{2022}, pp. \bibinfo{pages}{V--VII}.
\bibitem[{De~Raedt et~al.(2020)De~Raedt, Dumancic, Manhaeve, and Marra}]{de2020statistical}
\bibinfo{author}{L.~De~Raedt}, \bibinfo{author}{S.~Dumancic}, \bibinfo{author}{R.~Manhaeve}, \bibinfo{author}{G.~Marra},
\newblock \bibinfo{title}{From statistical relational to neuro-symbolic artificial intelligence},
\newblock in: \bibinfo{booktitle}{Proceedings of the Twenty-Ninth International Joint Conference on Artificial Intelligence, IJCAI 2020}, \bibinfo{organization}{ijcai. org}, \bibinfo{year}{2020}, pp. \bibinfo{pages}{4943--4950}.
\bibitem[{van Krieken et~al.(2022)van Krieken, Acar, and van Harmelen}]{van2022analyzing}
\bibinfo{author}{E.~van Krieken}, \bibinfo{author}{E.~Acar}, \bibinfo{author}{F.~van Harmelen},
\newblock \bibinfo{title}{Analyzing differentiable fuzzy logic operators},
\newblock \bibinfo{journal}{Artificial Intelligence} \bibinfo{volume}{302} (\bibinfo{year}{2022}) \bibinfo{pages}{103602}.
\bibitem[{Huang et~al.(2019)Huang, Liu, Zhai, Yin, Chen, Gao, and Hu}]{huang2019exploring}
\bibinfo{author}{Z.~Huang}, \bibinfo{author}{Q.~Liu}, \bibinfo{author}{C.~Zhai}, \bibinfo{author}{Y.~Yin}, \bibinfo{author}{E.~Chen}, \bibinfo{author}{W.~Gao}, \bibinfo{author}{G.~Hu},
\newblock \bibinfo{title}{Exploring multi-objective exercise recommendations in online education systems},
\newblock in: \bibinfo{booktitle}{Proceedings of the 28th ACM International Conference on Information and Knowledge Management}, \bibinfo{year}{2019}, pp. \bibinfo{pages}{1261--1270}.
\bibitem[{Koren(2008)}]{koren2008factorization}
\bibinfo{author}{Y.~Koren},
\newblock \bibinfo{title}{Factorization meets the neighborhood: a multifaceted collaborative filtering model},
\newblock in: \bibinfo{booktitle}{Proceedings of the 14th ACM SIGKDD international conference on Knowledge discovery and data mining}, \bibinfo{year}{2008}, pp. \bibinfo{pages}{426--434}.
\bibitem[{Adomavicius and Tuzhilin(2011)}]{adomavicius2011context}
\bibinfo{author}{G.~Adomavicius}, \bibinfo{author}{A.~Tuzhilin},
\newblock \bibinfo{title}{Context-aware recommender systems},
\newblock in: \bibinfo{booktitle}{Recommender systems handbook}, \bibinfo{publisher}{Springer}, \bibinfo{year}{2011}, pp. \bibinfo{pages}{217--253}.
\bibitem[{Karatzoglou et~al.(2010)Karatzoglou, Amatriain, Baltrunas, and Oliver}]{karatzoglou2010multiverse}
\bibinfo{author}{A.~Karatzoglou}, \bibinfo{author}{X.~Amatriain}, \bibinfo{author}{L.~Baltrunas}, \bibinfo{author}{N.~Oliver},
\newblock \bibinfo{title}{Multiverse recommendation: n-dimensional tensor factorization for context-aware collaborative filtering},
\newblock in: \bibinfo{booktitle}{Proceedings of the fourth ACM conference on Recommender systems}, \bibinfo{year}{2010}, pp. \bibinfo{pages}{79--86}.
\bibitem[{Koren(2009)}]{koren2009collaborative}
\bibinfo{author}{Y.~Koren},
\newblock \bibinfo{title}{Collaborative filtering with temporal dynamics},
\newblock in: \bibinfo{booktitle}{Proceedings of the 15th ACM SIGKDD international conference on Knowledge discovery and data mining}, \bibinfo{year}{2009}, pp. \bibinfo{pages}{447--456}.
\bibitem[{Zhang et~al.(2017)Zhang, Ai, Chen, and Croft}]{zhang2017joint}
\bibinfo{author}{Y.~Zhang}, \bibinfo{author}{Q.~Ai}, \bibinfo{author}{X.~Chen}, \bibinfo{author}{W.~B. Croft},
\newblock \bibinfo{title}{Joint representation learning for top-n recommendation with heterogeneous information sources},
\newblock in: \bibinfo{booktitle}{Proceedings of the 2017 ACM on Conference on Information and Knowledge Management}, \bibinfo{year}{2017}, pp. \bibinfo{pages}{1449--1458}.
\bibitem[{He and McAuley(2016)}]{he2016vbpr}
\bibinfo{author}{R.~He}, \bibinfo{author}{J.~McAuley},
\newblock \bibinfo{title}{Vbpr: visual bayesian personalized ranking from implicit feedback},
\newblock in: \bibinfo{booktitle}{Proceedings of the AAAI conference on artificial intelligence}, volume~\bibinfo{volume}{30}, \bibinfo{year}{2016}.
\bibitem[{Ai et~al.(2018)Ai, Azizi, Chen, and Zhang}]{ai2018learning}
\bibinfo{author}{Q.~Ai}, \bibinfo{author}{V.~Azizi}, \bibinfo{author}{X.~Chen}, \bibinfo{author}{Y.~Zhang},
\newblock \bibinfo{title}{Learning heterogeneous knowledge base embeddings for explainable recommendation},
\newblock \bibinfo{journal}{Algorithms} \bibinfo{volume}{11} (\bibinfo{year}{2018}) \bibinfo{pages}{137}.
\bibitem[{Hsieh et~al.(2017)Hsieh, Yang, Cui, Lin, Belongie, and Estrin}]{hsieh2017collaborative}
\bibinfo{author}{C.-K. Hsieh}, \bibinfo{author}{L.~Yang}, \bibinfo{author}{Y.~Cui}, \bibinfo{author}{T.-Y. Lin}, \bibinfo{author}{S.~Belongie}, \bibinfo{author}{D.~Estrin},
\newblock \bibinfo{title}{Collaborative metric learning},
\newblock in: \bibinfo{booktitle}{Proceedings of the 26th international conference on world wide web}, \bibinfo{year}{2017}, pp. \bibinfo{pages}{193--201}.
\bibitem[{Salakhutdinov et~al.(2007)Salakhutdinov, Mnih, and Hinton}]{salakhutdinov2007restricted}
\bibinfo{author}{R.~Salakhutdinov}, \bibinfo{author}{A.~Mnih}, \bibinfo{author}{G.~Hinton},
\newblock \bibinfo{title}{Restricted boltzmann machines for collaborative filtering},
\newblock in: \bibinfo{booktitle}{Proceedings of the 24th international conference on Machine learning}, \bibinfo{year}{2007}, pp. \bibinfo{pages}{791--798}.
\bibitem[{He et~al.(2017)He, Liao, Zhang, Nie, Hu, and Chua}]{he2017neural}
\bibinfo{author}{X.~He}, \bibinfo{author}{L.~Liao}, \bibinfo{author}{H.~Zhang}, \bibinfo{author}{L.~Nie}, \bibinfo{author}{X.~Hu}, \bibinfo{author}{T.-S. Chua},
\newblock \bibinfo{title}{Neural collaborative filtering},
\newblock in: \bibinfo{booktitle}{Proceedings of the 26th international conference on world wide web}, \bibinfo{year}{2017}, pp. \bibinfo{pages}{173--182}.
\bibitem[{Ebesu et~al.(2018)Ebesu, Shen, and Fang}]{ebesu2018collaborative}
\bibinfo{author}{T.~Ebesu}, \bibinfo{author}{B.~Shen}, \bibinfo{author}{Y.~Fang},
\newblock \bibinfo{title}{Collaborative memory network for recommendation systems},
\newblock in: \bibinfo{booktitle}{The 41st international ACM SIGIR conference on research \& development in information retrieval}, \bibinfo{year}{2018}, pp. \bibinfo{pages}{515--524}.
\bibitem[{Jiang et~al.(2017)Jiang, Zheng, Tan, Tang, and Zhou}]{jiang2017variational}
\bibinfo{author}{Z.~Jiang}, \bibinfo{author}{Y.~Zheng}, \bibinfo{author}{H.~Tan}, \bibinfo{author}{B.~Tang}, \bibinfo{author}{H.~Zhou},
\newblock \bibinfo{title}{Variational deep embedding: an unsupervised and generative approach to clustering},
\newblock in: \bibinfo{booktitle}{Proceedings of the 26th International Joint Conference on Artificial Intelligence}, \bibinfo{year}{2017}, pp. \bibinfo{pages}{1965--1972}.
\bibitem[{Hohenecker and Lukasiewicz(2020)}]{hohenecker2020ontology}
\bibinfo{author}{P.~Hohenecker}, \bibinfo{author}{T.~Lukasiewicz},
\newblock \bibinfo{title}{Ontology reasoning with deep neural networks},
\newblock \bibinfo{journal}{Journal of Artificial Intelligence Research} \bibinfo{volume}{68} (\bibinfo{year}{2020}) \bibinfo{pages}{503--540}.
\bibitem[{Makni and Hendler(2019)}]{makni2019deep}
\bibinfo{author}{B.~Makni}, \bibinfo{author}{J.~Hendler},
\newblock \bibinfo{title}{Deep learning for noise-tolerant rdfs reasoning},
\newblock \bibinfo{journal}{Semantic Web} \bibinfo{volume}{10} (\bibinfo{year}{2019}) \bibinfo{pages}{823--862}.
\bibitem[{Hinton et~al.(2006)Hinton, Osindero, and Teh}]{hinton2006fast}
\bibinfo{author}{G.~E. Hinton}, \bibinfo{author}{S.~Osindero}, \bibinfo{author}{Y.-W. Teh},
\newblock \bibinfo{title}{A fast learning algorithm for deep belief nets},
\newblock \bibinfo{journal}{Neural computation} \bibinfo{volume}{18} (\bibinfo{year}{2006}) \bibinfo{pages}{1527--1554}.
\bibitem[{Yang et~al.(2017)Yang, Yang, and Cohen}]{yang2017differentiable}
\bibinfo{author}{F.~Yang}, \bibinfo{author}{Z.~Yang}, \bibinfo{author}{W.~W. Cohen},
\newblock \bibinfo{title}{Differentiable learning of logical rules for knowledge base reasoning},
\newblock \bibinfo{journal}{Advances in neural information processing systems} \bibinfo{volume}{30} (\bibinfo{year}{2017}).
\bibitem[{Johnson et~al.(2017)Johnson, Hariharan, Van Der~Maaten, Hoffman, Fei-Fei, Lawrence~Zitnick, and Girshick}]{johnson2017inferring}
\bibinfo{author}{J.~Johnson}, \bibinfo{author}{B.~Hariharan}, \bibinfo{author}{L.~Van Der~Maaten}, \bibinfo{author}{J.~Hoffman}, \bibinfo{author}{L.~Fei-Fei}, \bibinfo{author}{C.~Lawrence~Zitnick}, \bibinfo{author}{R.~Girshick},
\newblock \bibinfo{title}{Inferring and executing programs for visual reasoning},
\newblock in: \bibinfo{booktitle}{Proceedings of the IEEE international conference on computer vision}, \bibinfo{year}{2017}, pp. \bibinfo{pages}{2989--2998}.
\bibitem[{Yi et~al.(2018)Yi, Wu, Gan, Torralba, Kohli, and Tenenbaum}]{yi2018neural}
\bibinfo{author}{K.~Yi}, \bibinfo{author}{J.~Wu}, \bibinfo{author}{C.~Gan}, \bibinfo{author}{A.~Torralba}, \bibinfo{author}{P.~Kohli}, \bibinfo{author}{J.~Tenenbaum},
\newblock \bibinfo{title}{Neural-symbolic vqa: Disentangling reasoning from vision and language understanding},
\newblock \bibinfo{journal}{Advances in neural information processing systems} \bibinfo{volume}{31} (\bibinfo{year}{2018}).
\bibitem[{Dong et~al.(2018)Dong, Mao, Lin, Wang, Li, and Zhou}]{dong2018neural}
\bibinfo{author}{H.~Dong}, \bibinfo{author}{J.~Mao}, \bibinfo{author}{T.~Lin}, \bibinfo{author}{C.~Wang}, \bibinfo{author}{L.~Li}, \bibinfo{author}{D.~Zhou},
\newblock \bibinfo{title}{Neural logic machines},
\newblock in: \bibinfo{booktitle}{International Conference on Learning Representations}, \bibinfo{year}{2018}.
\bibitem[{D{\'\i}az-Rodr{\'\i}guez et~al.(2022)D{\'\i}az-Rodr{\'\i}guez, Lamas, Sanchez, Franchi, Donadello, Tabik, Filliat, Cruz, Montes, and Herrera}]{diaz2022explainable}
\bibinfo{author}{N.~D{\'\i}az-Rodr{\'\i}guez}, \bibinfo{author}{A.~Lamas}, \bibinfo{author}{J.~Sanchez}, \bibinfo{author}{G.~Franchi}, \bibinfo{author}{I.~Donadello}, \bibinfo{author}{S.~Tabik}, \bibinfo{author}{D.~Filliat}, \bibinfo{author}{P.~Cruz}, \bibinfo{author}{R.~Montes}, \bibinfo{author}{F.~Herrera},
\newblock \bibinfo{title}{Explainable neural-symbolic learning (x-nesyl) methodology to fuse deep learning representations with expert knowledge graphs: The monumai cultural heritage use case},
\newblock \bibinfo{journal}{Information Fusion} \bibinfo{volume}{79} (\bibinfo{year}{2022}) \bibinfo{pages}{58--83}.
\bibitem[{Arabshahi et~al.(2021)Arabshahi, Lee, Gawarecki, Mazaitis, Azaria, and Mitchell}]{arabshahi2021conversational}
\bibinfo{author}{F.~Arabshahi}, \bibinfo{author}{J.~Lee}, \bibinfo{author}{M.~Gawarecki}, \bibinfo{author}{K.~Mazaitis}, \bibinfo{author}{A.~Azaria}, \bibinfo{author}{T.~Mitchell},
\newblock \bibinfo{title}{Conversational neuro-symbolic commonsense reasoning},
\newblock in: \bibinfo{booktitle}{Proceedings of the AAAI Conference on Artificial Intelligence}, volume~\bibinfo{volume}{35}, \bibinfo{year}{2021}, pp. \bibinfo{pages}{4902--4911}.
\bibitem[{Ma et~al.(2021)Ma, Nie, Yu, Jiang, Xiao, Zhu, Zhu, and Anandkumar}]{ma2021relvit}
\bibinfo{author}{X.~Ma}, \bibinfo{author}{W.~Nie}, \bibinfo{author}{Z.~Yu}, \bibinfo{author}{H.~Jiang}, \bibinfo{author}{C.~Xiao}, \bibinfo{author}{Y.~Zhu}, \bibinfo{author}{S.-C. Zhu}, \bibinfo{author}{A.~Anandkumar},
\newblock \bibinfo{title}{Relvit: Concept-guided vision transformer for visual relational reasoning},
\newblock in: \bibinfo{booktitle}{International Conference on Learning Representations}, \bibinfo{year}{2021}.
\bibitem[{Tsamoura et~al.(2021)Tsamoura, Hospedales, and Michael}]{tsamoura2021neural}
\bibinfo{author}{E.~Tsamoura}, \bibinfo{author}{T.~Hospedales}, \bibinfo{author}{L.~Michael},
\newblock \bibinfo{title}{Neural-symbolic integration: A compositional perspective},
\newblock in: \bibinfo{booktitle}{Proceedings of the AAAI Conference on Artificial Intelligence}, volume~\bibinfo{volume}{35}, \bibinfo{year}{2021}, pp. \bibinfo{pages}{5051--5060}.
\bibitem[{Lang and Poon(2021)}]{lang2021self}
\bibinfo{author}{H.~Lang}, \bibinfo{author}{H.~Poon},
\newblock \bibinfo{title}{Self-supervised self-supervision by combining deep learning and probabilistic logic},
\newblock in: \bibinfo{booktitle}{Proceedings of the AAAI Conference on Artificial Intelligence}, volume~\bibinfo{volume}{35}, \bibinfo{year}{2021}, pp. \bibinfo{pages}{4978--4986}.
\bibitem[{Amayuelas et~al.(2022)Amayuelas, Zhang, Rao, and Zhang}]{amayuelas2022neural}
\bibinfo{author}{A.~Amayuelas}, \bibinfo{author}{S.~Zhang}, \bibinfo{author}{S.~X. Rao}, \bibinfo{author}{C.~Zhang},
\newblock \bibinfo{title}{Neural methods for logical reasoning over knowledge graphs},
\newblock in: \bibinfo{booktitle}{International Conference on Learning Representations}, \bibinfo{year}{2022}.
\bibitem[{Shi et~al.(2020)Shi, Chen, Ma, Mao, Zhang, and Zhang}]{shi2020neural}
\bibinfo{author}{S.~Shi}, \bibinfo{author}{H.~Chen}, \bibinfo{author}{W.~Ma}, \bibinfo{author}{J.~Mao}, \bibinfo{author}{M.~Zhang}, \bibinfo{author}{Y.~Zhang},
\newblock \bibinfo{title}{Neural logic reasoning},
\newblock in: \bibinfo{booktitle}{Proceedings of the 29th ACM International Conference on Information \& Knowledge Management}, \bibinfo{year}{2020}, pp. \bibinfo{pages}{1365--1374}.
\bibitem[{Leshno et~al.(1993)Leshno, Lin, Pinkus, and Schocken}]{leshno1993multilayer}
\bibinfo{author}{M.~Leshno}, \bibinfo{author}{V.~Y. Lin}, \bibinfo{author}{A.~Pinkus}, \bibinfo{author}{S.~Schocken},
\newblock \bibinfo{title}{Multilayer feedforward networks with a nonpolynomial activation function can approximate any function},
\newblock \bibinfo{journal}{Neural networks} \bibinfo{volume}{6} (\bibinfo{year}{1993}) \bibinfo{pages}{861--867}.
\bibitem[{Harper and Konstan(2015)}]{harper2015movielens}
\bibinfo{author}{F.~M. Harper}, \bibinfo{author}{J.~A. Konstan},
\newblock \bibinfo{title}{The movielens datasets: History and context},
\newblock \bibinfo{journal}{Acm transactions on interactive intelligent systems (tiis)} \bibinfo{volume}{5} (\bibinfo{year}{2015}) \bibinfo{pages}{1--19}.
\bibitem[{Ni et~al.(2019)Ni, Li, and McAuley}]{ni2019justifying}
\bibinfo{author}{J.~Ni}, \bibinfo{author}{J.~Li}, \bibinfo{author}{J.~McAuley},
\newblock \bibinfo{title}{Justifying recommendations using distantly-labeled reviews and fine-grained aspects},
\newblock in: \bibinfo{booktitle}{Proceedings of the 2019 conference on empirical methods in natural language processing and the 9th international joint conference on natural language processing (EMNLP-IJCNLP)}, \bibinfo{year}{2019}, pp. \bibinfo{pages}{188--197}.
\bibitem[{Zhao et~al.(2020)Zhao, Chen, Wang, Gu, and Wen}]{zhao2020revisiting}
\bibinfo{author}{W.~X. Zhao}, \bibinfo{author}{J.~Chen}, \bibinfo{author}{P.~Wang}, \bibinfo{author}{Q.~Gu}, \bibinfo{author}{J.-R. Wen},
\newblock \bibinfo{title}{Revisiting alternative experimental settings for evaluating top-n item recommendation algorithms},
\newblock in: \bibinfo{booktitle}{Proceedings of the 29th ACM International Conference on Information \& Knowledge Management}, \bibinfo{year}{2020}, pp. \bibinfo{pages}{2329--2332}.
\bibitem[{Rendle et~al.(2012)Rendle, Freudenthaler, Gantner, and Schmidt-Thieme}]{rendle2012bpr}
\bibinfo{author}{S.~Rendle}, \bibinfo{author}{C.~Freudenthaler}, \bibinfo{author}{Z.~Gantner}, \bibinfo{author}{L.~Schmidt-Thieme},
\newblock \bibinfo{title}{Bpr: Bayesian personalized ranking from implicit feedback},
\newblock \bibinfo{journal}{arXiv preprint arXiv:1205.2618}  (\bibinfo{year}{2012}).
\bibitem[{Liu et~al.(2018)Liu, Zeng, Mokhosi, and Zhang}]{liu2018stamp}
\bibinfo{author}{Q.~Liu}, \bibinfo{author}{Y.~Zeng}, \bibinfo{author}{R.~Mokhosi}, \bibinfo{author}{H.~Zhang},
\newblock \bibinfo{title}{Stamp: short-term attention/memory priority model for session-based recommendation},
\newblock in: \bibinfo{booktitle}{Proceedings of the 24th ACM SIGKDD international conference on knowledge discovery \& data mining}, \bibinfo{year}{2018}, pp. \bibinfo{pages}{1831--1839}.
\bibitem[{Hidasi et~al.(2016)Hidasi, Karatzoglou, Baltrunas, and Tikk}]{hidasi2015session}
\bibinfo{author}{B.~Hidasi}, \bibinfo{author}{A.~Karatzoglou}, \bibinfo{author}{L.~Baltrunas}, \bibinfo{author}{D.~Tikk},
\newblock \bibinfo{title}{Session-based recommendations with recurrent neural networks},
\newblock in: \bibinfo{booktitle}{ICLR}, \bibinfo{year}{2016}.

\end{thebibliography}





\end{document}